\documentclass{article} 
\usepackage[table]{xcolor}
\usepackage{hyperref}
\usepackage{url}
\usepackage{graphicx}
\usepackage{booktabs}
\usepackage{multirow}
\usepackage{colortbl}
\usepackage{mathtools}
\usepackage{amsmath}
\usepackage{amssymb}
\usepackage{wrapfig}
\usepackage{pifont}
\usepackage{algorithm}
\usepackage{algorithmic}
\usepackage{amsthm}
\newcommand{\cmark}{\ding{51}}
\newcommand{\xmark}{\ding{55}}
\usepackage{natbib}

\usepackage{amsmath,amsfonts,bm}

\def\eqref#1{equation~\ref{#1}}

\def\1{\bm{1}}

\DeclareMathAlphabet{\mathsfit}{\encodingdefault}{\sfdefault}{m}{sl}
\SetMathAlphabet{\mathsfit}{bold}{\encodingdefault}{\sfdefault}{bx}{n}

\DeclareMathOperator{\Tr}{Tr}

\newtheorem{proposition}{Proposition}
\newtheorem{theorem}{Theorem}[section]
\newcommand{\epsz}{\varepsilon_{0}}             
\newcommand{\epsclip}{\varepsilon_{\mathrm{clip}}} 
\newcommand{\taus}{\tau_{s}}                    
\newcommand{\Vocab}{\mathcal{V}}                
\theoremstyle{plain}

\theoremstyle{definition}

\theoremstyle{remark}
\newtheorem{remark}[theorem]{Remark}

\title{DEEPO: Dual-Entropy Enhanced Policy Optimization for Hallucination in MLLMs}

\author{
Yingxuan Zhuang$^{1}$ \quad
Miao Pan$^{1}$ \quad
Wangjie Gan$^{1}$ \quad
Jingxiao Yang$^{1}$ \\
Fan Wang$^{1}$ \quad
Weiming Liu$^{1}$ \quad
Cheng Tan$^{2}$\quad
Xuhong Zhang$^{1}$\\
Jintao Chen$^{1}$\\
$^{1}$Zhejiang University \\
$^{2}$Shanghai Artificial Intelligence Laboratory \\
}

\begin{document}

\date{}
\maketitle

\begin{abstract}

Reinforcement learning (RL) is widely used to sharpen reasoning in multimodal large language models (MLLMs), yet its effect on hallucination is uneven. We trace this to two weak points in the \emph{correction chain} from reward to parameter update. At the rollout level, hard queries---those with high semantic entropy---frequently produce unanimously wrong sample groups, collapsing the group-relative
  advantage to zero exactly where hallucination risk is highest. At the optimization level, confident-but-wrong tokens are gradient-invisible: a categorical policy's expected score-gradient norm vanishes as its distribution sharpens, so the predictions that most need correction receive the weakest updates. We propose Dual-Entropy Enhanced Policy Optimization (DEEPO), a dual-stage enhancement combining signal
  variance regularization with gradient preconditioning: semantic-entropy-triggered expert prefixes inject grounded continuations on high-uncertainty queries, providing direct supervision and restoring advantage variance, while advantage-sign-aware R\'enyi preconditioning counteracts logit-level saturation so correction reaches confident errors in the operational confidence regime. Both branches improve over GRPO individually; their interaction is statistically significant on VideoMMMU---the most complex long-horizon task in our evaluation suite ($+4.0$, 95\% CI $[1.1, 6.9]$)---and additive elsewhere. DEEPO reduces hallucination while preserving accuracy and training stability.
\end{abstract}

\section{Introduction}
\label{sec:intro}

Multimodal large language models (MLLMs) have greatly advanced vision--language understanding and multi-step reasoning \citep{yin2024survey,wu2024visionllm,li2024enhancing}. However, hallucination---confident but incorrect or semantically inconsistent outputs---remains a key obstacle \citep{fu2025mitigating}, motivating work on its causes and mitigation \citep{mahmoud2025unintended}.

While prior work attributes hallucination to dataset noise \citep{bang2025hallulens}, architectural bottlenecks \citep{cossio2025comprehensive}, or decoding errors \citep{wu2025improve}, these explanations target static properties of the model or input \citep{dang2025survey}.
Hallucination is also shaped during training: SFT overfits surface patterns by imitation, whereas RL can encourage guessing under uncertainty through reward-driven rollouts~\citep{lin2024flame,mahmoud2025unintended,shenfeld2025rl}---an effect that is heterogeneous rather than universal (in our measurements GRPO improves over its base model on POPE and HallusionBench yet slightly degrades on VideoHallucer; Table~\ref{tab:chair}): the issue is not that RL uniformly worsens hallucination, but that its corrective machinery is unreliable precisely on confident errors.
We therefore focus on the RL stage, where the correction chain from scalar reward to parameter update has two weak points that a training method can, and we argue should, address jointly.

\begin{figure*}[htbp]
    \centering
    \includegraphics[width=\linewidth]{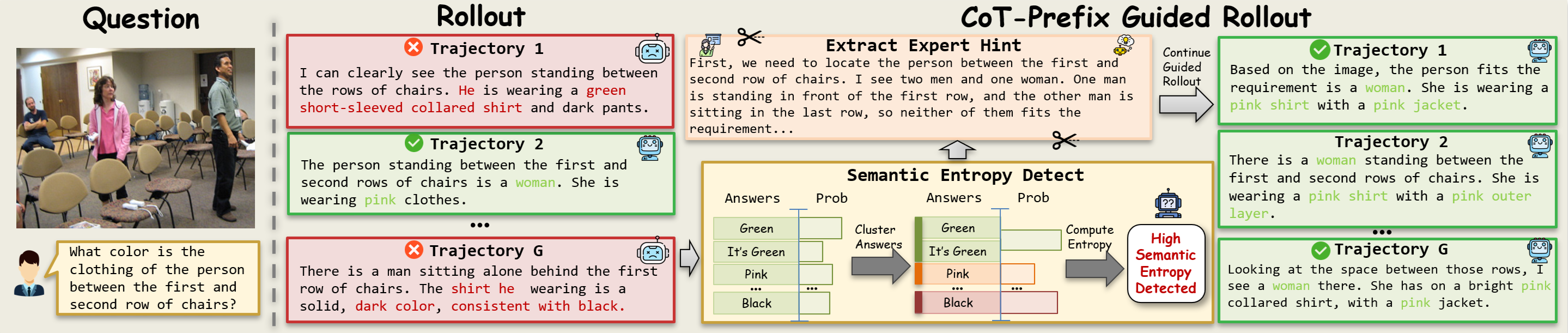}
      \caption{Model guessing and expert-hint-guided correction. Under uncertain visual cues the model ``guesses,'' yielding high semantic entropy and frequently unanimous failures; the expert hint grounds one continuation on the visual evidence.}
    \label{fig:Intro_Example}
\end{figure*}

\textbf{(I) Hard queries starve the correction signal.}
In Fig~\ref{fig:Intro_Example}, hallucinated answers concentrate in regions of high semantic entropy---exactly the queries on which sampling is most likely to fail unanimously: with binary rewards an all-wrong group has $\mu = 0$, $\sigma \to 0$, so every response receives $A_i \approx 0$ and the iteration extracts no gradient~\citep{kalai2025language,khan2024consistency}.
An \emph{expert hint}---a truncated, answer-free chain-of-thought (CoT) prefix---conditions one continuation on grounded visual reasoning, injecting direct supervision on hard queries and, under a shared group baseline, a high-reward anchor that restores nonzero advantage variance, so failed responses receive genuine negative advantages instead of the degenerate zero; prefixes come from dataset traces~\citep{videor12025} or the policy's own verified rollouts (Sec.~\ref{sec:expert_hint}), never a test-time oracle.

\textbf{(II) Confident errors are gradient-invisible.}
Even a perfect advantage signal cannot correct confident hallucinations, because the delivery mechanism vanishes: for a softmax policy, $\mathbb{E}\lVert\nabla_z\log\pi\rVert^2 = 1-e^{-H_2}$ (Prop.~\ref{th:relation}), and a gradient-masking pilot confirms that high-entropy tokens carry the bulk of the update signal (Fig.~\ref{fig:driver2-grad}, App.~\ref{app:additional_experiments}).
Confident predictions---and confident hallucinations in particular, which are precisely the errors that survive decoding---thus receive vanishing corrective updates, while uncertain regions dominate the gradient budget.

\textbf{Why the two stages are complementary.}
Rescued advantage variance is of little use if correction cannot reach confident errors, and curvature-compensated updates cannot manufacture signal from a collapsed group. DEEPO therefore combines the two: semantic entropy detects the queries whose groups are likely to collapse and triggers grounded continuations, while sign-aware Rényi entropy parameterizes how strongly each token is corrected or reinforced---signals at different levels, not interchangeable (App.~\ref{app:entropy_relation}). The interaction is positive on all six benchmarks and statistically significant exactly where both mechanisms are most stressed---long-horizon VideoMMMU ($+4.0$, 95\% CI $[1.1, 6.9]$; Table~\ref{tab:wo_POPE})---and consistent with additivity elsewhere.

\noindent\textbf{Our contributions are summarized as follows:}
\begin{itemize}
    \item \textbf{Diagnosis.} Two weak points in the RL correction chain: on hard queries, unanimously wrong groups collapse the group-relative advantage to zero, and confident errors are gradient-invisible (Prop.~\ref{th:relation}).
    \item \textbf{Dual-stage enhancement.} DEEPO pairs entropy-triggered expert prefixes at the rollout stage with advantage-sign-aware R\'enyi gradient preconditioning at the optimization stage; the scaling weight is a conservative damped proximal step with guaranteed ascent of a sample-wise surrogate (Thm.~\ref{thm:trace_asym}), and sign-symmetric rules provably cannot realize its allocation (Remark~\ref{rem:signsym}).
    \item \textbf{Evidence.} Complementary single-branch gains with a positive interaction---significant on VideoMMMU, additive elsewhere---supported by design, fairness, and mechanism-diagnosis controls, a self-generated-hint variant, multi-seed evaluation, and efficiency accounting.
\end{itemize}

\section{Related Work}
\label{sec:related}

\noindent\textbf{Hallucination mitigation in MLLMs.}
Hallucination remains a persistent challenge for MLLMs~\cite{bai2024hallucination}, addressed through attention calibration~\cite{xu2025mitigating, objectdpo2025}, decoding-time corrections~\cite{leng2024vcd, sun2024mvp}, grounding supervision~\cite{zhao2025marine, chen2023visualsupervision}, and multi-object or cross-image analyses~\cite{zhou2023lure, chen2024multiobject, park2025second, li2025mihbench}.
These are post-hoc or decoding-level fixes; we instead detect hallucination-prone uncertainty during rollout and ground the generation.

\noindent\textbf{Training dynamics, rewards, and uncertainty.}
RL can exacerbate hallucinations when rewards are weakly tied to factual correctness~\cite{he2024topic, yang2025opadpo} or encourage shortcut exploitation~\cite{miao2025energyloss, turn0academia20}.
Token-level credit assignment has been approached via perception-aware rewards~\cite{papo2025, vppo2025} and verifiable-data synthesis~\cite{synthrl2025}; these modify rewards or data, whereas DEEPO leaves the reward unchanged and modifies rollout conditioning and gradient delivery.
Uncertainty signals have mainly served detection or calibration~\cite{farquhar2024detecting, manakul2023selfcheckgpt, fadeeva2024tokenlevel, zhang2024calibrating, qiu2024semanticdensity, niu2024uq4ct, jiang2024cape,yang2026reconcilingprocesssupervisionoutcomebased,zhuang2026mitigatingmanifolddepartureuncertaintyaware}, not for controlling how uncertainty shapes rollouts and updates during RL fine-tuning.

\noindent\textbf{Positioning against three adjacent lines.}
\emph{(i) Entropy-modulated policy gradients} (EMPG~\cite{wang2025harnessing} and relatives) reshape token updates via uncertainty but are sign-symmetric (Remark~\ref{rem:signsym}) with no rollout-time mechanism.
\emph{(ii) Hallucination-targeted preference optimization} (RLHF-V~\cite{rlhfv2024}, RLAIF-V~\cite{rlaifv2024}, HA-DPO~\cite{hadpo2024}, hallucination-targeted DPO~\cite{fu2025mitigating}, SymMPO~\cite{sympo2025}, POVID~\cite{povid2024}) acts on offline preference pairs with no generation-time intervention; DEEPO is dynamics-centric and on-policy.
\emph{(iii) Expert-guided RL}~\cite{zhuang2026dualaxispolicyoptimizationllm,zhang2025policy,srft2025,yang2026saresamplewiseadaptivereasoning} mixes complete off-policy expert traces into on-policy RL at a static ratio---we adopt the confidence weight $\phi_t$ of \cite{zhang2025policy}---whereas DEEPO uses expert text only as truncated, answer-free prefixes, triggered per query by an adaptive semantic-entropy threshold, with hint dropout and the prefix masked out of the RL loss.
To our knowledge, DEEPO is the first to pair answer-level semantic entropy as a rollout-time trigger for expert conditioning with a token-level sign-aware R\'enyi scaling law.

\section{Preliminaries}
\label{sec:preliminaries}

\paragraph{GRPO.}
We build on Group Relative Policy Optimization (GRPO)~\citep{grpoorigin2024}, a critic-free RL algorithm: for each query $q$ the policy samples $G$ responses with verifiable rewards $r_i$ (task accuracy plus format compliance, following~\cite{videor12025}); the group-relative advantage $A_i = (r_i - \mathrm{mean})/\mathrm{std}$ is broadcast to every token, and the policy is updated with a PPO-style clipped surrogate plus a KL penalty to a frozen reference.
Two properties matter: (i) advantages are \emph{sequence-level}---the signal cannot distinguish which tokens caused failure; (ii) with binary rewards a unanimous group carries no signal---if all responses are wrong ($r_i = 0$) then $\mu = 0$, $\sigma \to 0$, and $A_i \approx 0$ (symmetrically for an all-correct group), so the iteration extracts no gradient precisely from the hardest queries.

\paragraph{Semantic entropy.}
Following~\cite{farquhar2024detecting}, semantic entropy measures guessing at the level of meaning: draw $N$ generations, partition them into $K$ meaning-equivalent clusters $\{C_k\}$, and compute $H_s(q) = -\sum_{k} P(C_k \mid q)\log P(C_k \mid q)$ over the cluster distribution; a high $H_s(q)$ indicates semantic-level guessing and elevated hallucination risk.
We estimate $H_s$ from the $G$ rollouts already sampled for the current iteration ($N{=}G$ by default; App.~\ref{app:hyperparameter_sensitivity} co-varies both), so estimation adds no extra sampling.

\paragraph{R\'enyi entropy.}
At order $\alpha = 2$, the collision entropy $H_2 = -\log\sum_y \pi(y\mid q,y_{<t})^2$ characterizes token-level sharpness and governs the expected score-gradient norm (Prop.~\ref{th:relation}); it is the quantity we use for gradient scaling.

\section{Method}
\label{sec:method}
\begin{figure*}[htbp]
    \centering
    \includegraphics[width=\linewidth]{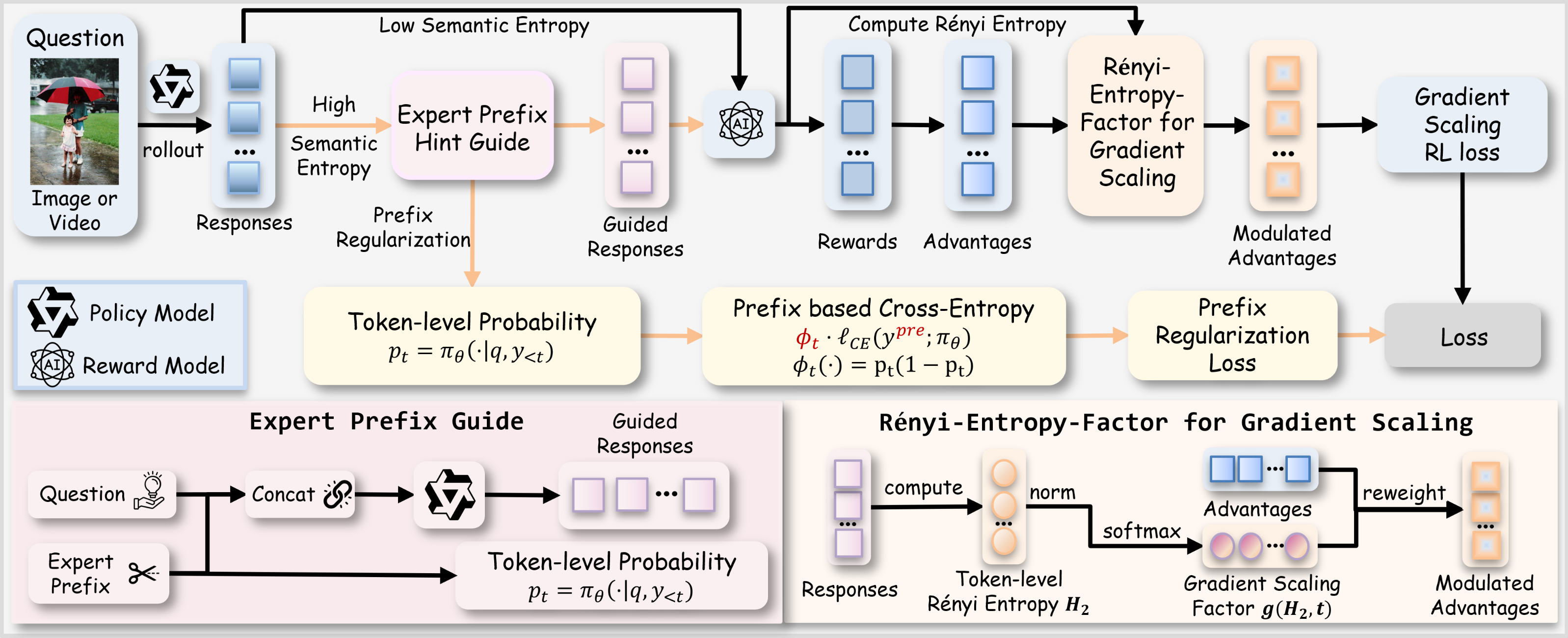}
    \caption{Overview of Dual-Entropy Enhanced Policy Optimization. High semantic entropy activates $\phi_t$-weighted prefix regularization and prefix-guided rollouts, while the R\'enyi entropy factor scales policy gradients; the objective unifies both.}
    \label{fig:Overview}
\end{figure*}

DEEPO addresses the two weak points identified in Sec.~\ref{sec:intro} as a dual-stage enhancement (Fig.~\ref{fig:Overview}): Sec.~\ref{sec:expert_hint} gives the rollout stage, where semantic-entropy-triggered expert prefixes provide grounded supervision and reshape in-group advantages on hard queries (signal repair); Sec.~\ref{sec:renyi_scaling} gives the optimization stage, where advantage-sign-aware R\'enyi gradient preconditioning delivers correction to confident tokens (delivery repair); Sec.~\ref{sec:coupling} states the complementarity hypothesis and its falsifiable predictions. Appendix~\ref{app:entropy_relation} relates the two entropy quantities through predictive uncertainty.

\vspace{-1em}

\subsection{Expert Hint as Signal Repair}\label{sec:expert_hint}

The first weak point in the correction chain is a starved learning signal on hard queries: when semantic entropy is high the sampled group fails frequently, and in the extreme unanimously ($r_i = 0$ for all $i$, hence $\mu = 0$, $\sigma \to 0$, and $A_i = (r_i - \mu)/(\sigma + \epsilon) \approx 0$ for every token), so no gradient is extracted exactly where hallucination risk is highest.
Entropy-triggered expert prefixes counteract this through two channels, separated explicitly because the evidence weighs them differently (Sec.~\ref{sec:causal_controls}): \emph{direct grounding and supervision} (the prefix steers the policy onto a grounded reasoning path, reinforced by RL and distilled through $\mathcal{L}_{\mathrm{aux}}$, Eq.~\ref{eq:aux_loss}) and \emph{in-group advantage reshaping} (under a shared group baseline, the high-reward grounded continuation raises the group mean, so unhinted failures receive real negative advantages instead of the degenerate $A_i \approx 0$ of a collapsed group).

\paragraph{Hint source.}
Hints are truncated prefixes of model-generated solution trajectories from the Video-R1 dataset~\citep{videor12025}; they condition generation during training only (never at inference) and additionally receive the lightweight $\phi_t$-weighted cross-entropy of Eq.~\ref{eq:aux_loss}, so DEEPO is a hybrid of RL on self-generated tokens and auxiliary prefix supervision, ablated separately (Table~\ref{tab:hint_control}, App.~\ref{app:controls}; Table~\ref{tab:wo_POPE}).
The method is not tied to this dataset: Sec.~\ref{sec:causal_controls} evaluates a self-generated variant (\textsc{Deepo}-self) using the prefix of the best verified-correct rollout from the same group.

\paragraph{Trigger and prefix construction.}
We estimate $H_s$ from the $N$ rollouts already generated for the current iteration (default $N{=}G$), clustering answer-level commitments by pairwise NLI entailment (DeBERTa-MNLI; App.~\ref{app:training_details}).
A query is marked as challenging when $H_s(q)$ exceeds an adaptive threshold $\hat{\tau}$, initialized at $\taus$ (default $0.8$) and updated per iteration by exponential moving average, $\hat{\tau} \leftarrow (1-\beta)\hat{\tau} + \beta\overline{H}_s$ with $\overline{H}_s$ the batch mean; the trigger gates the intervention to queries whose groups fail frequently, unlike a rate-matched random trigger (Table~\ref{tab:design_controls}).
Each expert trace is split at the final-answer marker into a thought segment $y^{*}_{\mathrm{thought}}$ (observations and chain-of-thought) and an answer segment $y^{*}_{\mathrm{ans}}$ (conclusion and options); the prefix is drawn \emph{exclusively from the thought segment}, with uncertainty-proportional length hard-capped away from the answer boundary:
\begin{equation}
\label{eq:prefix_len}
L_h = \min\Big(\big\lfloor \alpha_{\max} \cdot \tfrac{H_s(q)}{H_s^{\max}} \cdot L_{\mathrm{thought}} \big\rfloor,\; L_{\mathrm{thought}} - 1 \Big),
\end{equation}
where $H_s^{\max} = \log N$, $L_{\mathrm{thought}} = |y^{*}_{\mathrm{thought}}|$, and $\alpha_{\max} \in (0,1)$ (default $0.5$).
Since $\alpha_{\max} < 1$ and $H_s(q) \le H_s^{\max}$, $L_h \le L_{\mathrm{thought}} - 1$ for every trace: \textbf{the prefix provably excludes the final answer and conclusion tokens, and an NLI entailment filter additionally discards any prefix that semantically entails the gold answer (App.~\ref{app:training_details}).}
The policy then generates the continuation $y_q^{\mathrm{rem}} \sim \pi_\theta(\cdot \mid q \oplus y_q^{\mathrm{pre}})$; the full trajectory is scored by the verifiable reward (one extra generation pass for triggered queries; cost in Table~\ref{tab:efficiency}, App.~\ref{app:training_details}).

\paragraph{Two channels of benefit.}
For a triggered query, the hint-conditioned continuation and the first-pass unhinted rollouts \emph{share a single group-relative baseline}, and the prefix additionally receives $\mathcal{L}_{\mathrm{aux}}$.
\emph{(i) Direct grounding and supervision}: the continuation is generated by the policy itself and reinforced when successful, while $\mathcal{L}_{\mathrm{aux}}$ distills the prefix tokens; this channel operates whenever a grounded trace is available---including the policy's own correct rollouts (\textsc{Deepo}-self)---and does not depend on group composition.
\emph{(ii) In-group advantage reshaping}: under the shared baseline the high-reward continuation raises $\mu$ and restores advantage variance, so on a query whose unaided group would have failed unanimously, the failed unhinted rollouts---frozen at zero advantage---now receive substantial negative advantages; this channel punishes failures with real updates and by construction cannot separate two equally rewarded responses, and it requires the shared baseline (Table~\ref{tab:mechanism_diag}(c)).
Hint dropout regularly reverts triggered queries to purely unhinted groups.

\paragraph{Empirical diagnosis of advantage collapse.}
On the training distribution, the fraction of all-wrong groups rises monotonically from $12\%$ (lowest $H_s$ decile) to $63\%$ (highest), and on triggered queries whose first-pass group was unanimously wrong ($48\%$ of triggered queries), hint activation converts $60\%$ to mixed groups, moving the mean advantage of failed unhinted rollouts from $\approx 0$ to $-0.212$ and the within-group advantage variance from $0$ to $0.6$ (Table~\ref{tab:mechanism_diag}, Sec.~\ref{sec:controls}).
One coverage boundary is worth stating plainly: a policy that is confidently and \emph{consistently} wrong produces \emph{low} semantic entropy and is not triggered; and if such a group is also unanimously wrong, its advantage vanishes, so the optimization stage---which can only reallocate a nonzero advantage, since $g^\star(H_{2,t}, A_t)\, A_t = 0$ when $A_t = 0$---has no signal to amplify either.
Neither branch manufactures reward information the rollouts do not contain; this boundary is shared by every outcome-reward RL method, and the delivery branch acts on confident errors inside groups that carry nonzero advantage variance.

\paragraph{Optimization treatment of the prefix.}
The prefix is masked out of the RL loss (policy gradients update only self-generated tokens), and the guidance is distilled by an auxiliary prefix loss with confidence weight $\phi_t = \pi_\theta(y^{\mathrm{pre}}_{q,t}\mid q, y^{\mathrm{pre}}_{q,<t})\,(1-\pi_\theta(y^{\mathrm{pre}}_{q,t}\mid q, y^{\mathrm{pre}}_{q,<t}))$~\citep{zhang2025policy}:
\begin{equation}
\label{eq:aux_loss}
\mathcal{L}_{\mathrm{aux}} = \mathbb{E}_{(q,\,y^{\mathrm{sol}})\sim\mathcal{D}}\Big[\tfrac{\mathbb{1}\{L_h>0\}}{L_h}\textstyle\sum_{t=1}^{L_h}
\phi_t \, \ell_{\mathrm{CE}}\big(y^{\mathrm{pre}}_{q,t};\, \pi_\theta(\cdot \mid q, y^{\mathrm{pre}}_{q,<t})\big)\Big],
\end{equation}
where $\phi_t$ concentrates learning on tokens of intermediate confidence; $\mathcal{L}_{\mathrm{aux}}$ is a weak regularizer and leaves the RL objective unchanged.

\subsection{R\'enyi Gradient Scaling as Delivery Repair}
\label{sec:renyi_scaling}

A diagnostic signal has limited effect if it cannot reach the parameters responsible for confident hallucinations.
We now make the delivery bottleneck precise and derive the scaling rule, starting from the second-order R\'enyi entropy of the token-level policy, $H_2(\pi(\cdot \mid q, y_{<t})) = -\log \sum_{y \in \Vocab} \pi(y \mid q, y_{<t})^2$, which for softmax policies pins the expected magnitude of the score-function gradient:

\begin{proposition}[Score-gradient norm via collision entropy; standard identity]
\label{th:relation}
For a policy $\pi_\theta$ parameterized by logits $z_\theta(q,y_{<t})$,
\begin{equation}\label{eq:gradient_entropy_relation}
\mathbb{E}_{y_t \sim \pi_\theta(\cdot |q,y_{<t})}
\big[\big\lVert \nabla_{z_\theta(q,y_{<t})} \log \pi_\theta(y_t \mid q,y_{<t}) \big\rVert^2 \big]
= 1 - \exp\big(- H_2(\pi_\theta)\big).
\end{equation}
\end{proposition}

This is the trace of the categorical Fisher information in natural parameters (proof in App.~\ref{proof_entropy_gradient}): hallucinated outputs are often \emph{confidently wrong}, and by Eq.~\ref{eq:gradient_entropy_relation} such low-entropy predictions induce near-zero score gradients, so RL provides the weakest correction exactly where it is most needed.

\paragraph{From the identity to a weighting rule.}
EMPG~\citep{wang2025harnessing} reweights token updates with an entropy-only factor, which is symmetric in update direction.
We instead derive a sign-aware weight from a \emph{sample-wise} proximal analysis in logit space, conditioned on the current batch of \emph{already generated} trajectories: every $A_t$ is a fixed empirical scalar, and no expectation over $y_t \sim \pi_\theta$ is taken with $A_t$ inside---the only distribution-level quantity is the Fisher matrix $F_t = \operatorname{diag}(\pi) - \pi\pi^\top \succeq 0$, a deterministic function of the policy at the current state.
Defining
\begin{equation}
\label{eq:tau_def}
\tau_t := \Tr(F_t) = 1 - \exp(-H_{2,t}),
\end{equation}
the following theorem (proved in App.~\ref{app:proof_trace_asym}) justifies a curvature-compensated weight under softmax saturation; the subsequent remark states what entropy-only rules cannot do.

\begin{theorem}[Curvature-compensated weight as a conservative damped proximal step]
\label{thm:trace_asym}
Fix a generated batch (hence all advantages $\{A_t\}$ are fixed empirical scalars, $A_t \neq 0$) and a decoding step $t$ with state $s_t = (q, y_{<t})$, sampled token $y_t$, and score gradient $g_t = \nabla_{z_t}\log\pi_\theta(y_t \mid s_t)$.
Restrict one-step logit updates to the score direction, $\Delta z_t = \eta\, w_t A_t g_t$ with $w_t \ge 0$, and consider the sample-wise second-order proximal surrogate with Levenberg--Marquardt damping $\epsz > 0$,
$Q_t(w_t) = A_t\, g_t^\top \Delta z_t - \tfrac{\beta_{\mathrm{kl}}}{2}\, \Delta z_t^\top (F_t + \epsz I)\, \Delta z_t$.
Then:
\begin{itemize}
    \item[(i)] \emph{Sample-wise curvature bound.} $g_t^\top F_t g_t \le \tau_t \lVert g_t \rVert^2$, so the trace-scalarized quadratic
    $\widetilde Q_t(w) = \eta A_t^2 \lVert g_t \rVert^2\, w - \tfrac{\beta_{\mathrm{kl}}\eta^2}{2} A_t^2 \lVert g_t \rVert^2 (\tau_t + \epsz)\, w^2$
    is a tangent minorant of $Q_t$ on $\mathbb{R}_+$.
    \item[(ii)] \emph{Damped optimum.} The minorant maximizer
    $w_t^{\epsz} = \big(\beta_{\mathrm{kl}}\,\eta\,(\tau_t + \epsz)\big)^{-1} \le (\beta_{\mathrm{kl}}\,\eta\,\epsz)^{-1} =: w_{\max}$
    is capped and guarantees strict sample-wise ascent
    $Q_t(w_t^{\epsz}) \ge A_t^2\lVert g_t\rVert^2 / \big(2\beta_{\mathrm{kl}}(\tau_t + \epsz)\big) > 0 = Q_t(0)$.
    \item[(iii)] \emph{Softmax-saturation regime.} With $\delta_t = 1 - \pi_\theta(y_t \mid s_t)$, $\lVert g_t \rVert^2 \le 2\delta_t^2$ and $\tau_t \le 2\delta_t$: as $\delta_t \to 0$ the unweighted step stalls ($Q_t'(0) = O(\delta_t^2)$), the compensated weight grows as $\Omega(\delta_t^{-1})$ up to its cap, and the certified ascent decays only as $\Theta(\delta_t)$ for $\delta_t \gg \epsz$; in the deep tail $\delta_t \ll \epsz$ the bound reverts to $O(\delta_t^2/\epsz)$---a rate improvement over the operational confidence band, not an asymptotic removal of saturation.
\end{itemize}
\end{theorem}

Three readings are essential: the statement is \emph{sample-wise and local} ($A_t$ never appears inside an expectation over $y_t \sim \pi_\theta$; one step at one token of a fixed trajectory, not the global return); the damping is \emph{load-bearing} (the $+\epsz$ term, exactly the stabilizer in Eq.~\ref{eq:phi_rule}, caps the weight at $w_{\max}$); and the theorem is \emph{sign-blind} ($Q_t$ sees $A_t$ only through $A_t^2$), justifying the \emph{magnitude} of the corrective weight, not its sign asymmetry.

\begin{remark}[Entropy-only rules cannot prioritize correction]
\label{rem:signsym}
If $w_t = \nu(H_{2,t})$ is any advantage-sign-independent rule, two tokens with $H_{2,a} = H_{2,b}$ and $A_a < 0 < A_b$ necessarily get $w_a = w_b$, so no such rule can implement correction-priority allocation $w_a > w_b$; under any fixed batch budget, a budget-preserving perturbation strictly increases the corrective mass (App.~\ref{app:proof_trace_asym}).
The division of labor is explicit: Theorem~\ref{thm:trace_asym} provides a magnitude justification for inverse-curvature weighting under a local quadratic surrogate, while the sign-asymmetric assignment is an empirical heuristic for correction priority, validated by the weighting-rule ablation (Table~\ref{tab:design_controls}).
\end{remark}

Guided by Theorem~\ref{thm:trace_asym} and Remark~\ref{rem:signsym}, we apply a sign-aware R\'enyi factor to the advantage:
\begin{equation}
\label{eq:phi_rule}
g^{\star}(H_{2,t},A_t)= \psi(H_{2,t},A_t)\Big/\bar\psi,\qquad
\psi(H_{2,t},A_t)=
\begin{cases}
\bigl(\tau_t + \epsz\bigr)^{-1}, & A_t<0 \quad\text{(correction)},\\[2pt]
\tau_t, & A_t\ge 0 \quad\text{(reinforcement)},
\end{cases}
\end{equation}
where $\bar\psi$ is the batch mean over all tokens, $\tau_t = 1 - e^{-H_{2,t}}$, and $\epsz = 1/w_{\max}$ with weight cap $w_{\max} = 20$ (App.~\ref{app:training_details}).
The negative branch is the damped proximal weight of Theorem~\ref{thm:trace_asym}(ii) with $\beta_{\mathrm{kl}},\eta$ absorbed by batch normalization.
Its effect is best read through \emph{effective step magnitude}, not token selection: a confident token has $\lVert g_t\rVert = O(\delta_t)$, so its unweighted logit step collapses as the prediction saturates, and weighting by $(\tau_t+\epsz)^{-1} \approx (2\delta_t+\epsz)^{-1}$ recovers the effective step $\lVert\Delta z_t\rVert \propto w_t\lVert g_t\rVert$ to a $\delta_t$-independent scale wherever $\delta_t \gg \epsz$ (Thm.~\ref{thm:trace_asym}(iii)).
The rule thus un-stalls optimization on confident errors within the operational confidence regime; it does not remove saturation asymptotically (at fixed $\epsz > 0$ the weight saturates at $w_{\max}$ and the step still vanishes as $\delta_t \to 0$).
This amplification is \emph{not} error-selective by construction---low-entropy structural tokens (articles, connectives) in failed trajectories are amplified as well---but three properties bound that collateral: the cap at $w_{\max}$ keeps amplified steps proportional to an already-small $\lVert g_t\rVert$; the KL penalty to the frozen reference is deliberately left unweighted (Sec.~\ref{sec:coupling}), anchoring high-frequency tokens; and $\mathcal{L}_{\mathrm{aux}}$ adds an orthogonal cross-entropy pull on fluent text.
The residual selectivity is empirical: on span-annotated failed rollouts the updates achieve a $1.75$--$1.80\times$ per-token \emph{weight} enrichment on hallucinated spans (95\% CIs exclude $1$; App.~\ref{app:further_controls}), i.e.\ a $1.15$--$1.25\times$ enrichment of the effective logit-step norm---a deliberately bounded effect: restoring a stalled gradient by roughly a fifth suffices, over thousands of updates, to bias the policy away from suboptimal deterministic attractors.
The positive branch is a separate, deliberately simple heuristic: reinforce in proportion to the available gradient signal $\mathbb{E}\lVert g\rVert^2 = \tau_t$, dampening reinforcement of confident tokens while leaving high-uncertainty tokens near the unweighted baseline; confident correct tokens receive small positive weights, slowing further sharpening and acting against entropy collapse.
Batch-mean normalization keeps the weights centered at one, so the rule reallocates update magnitude across tokens rather than rescaling the batch mean; $g^\star > 0$ preserves the advantage sign, so the PPO clip \emph{trigger set} is unchanged.

\paragraph{Interaction with downstream optimization (Adam and PPO).}
Theorem~\ref{thm:trace_asym} characterizes a single damped proximal step; the deployed pipeline composes it with Adam and PPO clipping, which the theorem does not model.
Two observations keep the design meaningful: $g^\star$ acts in the \emph{spatial} dimension (per token, within a batch) whereas Adam's normalization acts \emph{temporally} (running statistics across steps), rescaling each token's update by roughly a common factor so the relative, mean-one enrichment survives and steers subsequent updates; and $g^\star > 0$ leaves the clip trigger set unchanged, so clipping intervenes on trajectories, not on the weighting.
We therefore read the theorem as a design-level preconditioning argument, not a guarantee about the composed optimizer (details in App.~\ref{app:training_details}).

\subsection{The Unified Objective}
\label{sec:coupling}

\paragraph{Unified objective.}
The final training loss (minimized) combines the GRPO surrogate on self-generated continuations with the weighted prefix regularizer:
\begin{multline}\label{eq:grpo_total}
\mathcal{L} = \mathbb{E}_{\mathcal{D}}\Big[
\lambda\, \mathcal{L}_{\mathrm{aux}}
- \tfrac{1}{|y_q^{\mathrm{rem}}|}\textstyle\sum_{t=1}^{|y_q^{\mathrm{rem}}|}
\min\big(r_t \tilde A_t, \mathrm{clip}(r_t, 1{-}\epsclip, 1{+}\epsclip) \tilde A_t\big)\Big]
+ \beta_{\mathrm{kl}}\, \mathbb{E}\, D_{\mathrm{KL}}(\pi_\theta \| \pi_{\mathrm{ref}}),
\end{multline}
where $r_t$ is the policy ratio, $\tilde A_t = g^\star(H_{2,t}, A_t)\cdot A_t$ the scaled advantage, and $\beta_{\mathrm{kl}} = 0.04$; untriggered queries have $L_h = 0$.
The KL term is deliberately \emph{not} scaled by $g^\star$: correction tokens are allowed to move further from the reference policy (App.~\ref{app:proof_trace_asym}).

\paragraph{Complementarity hypothesis.}
If the two stages repair distinct weak points, then (P1) grounded hints should help while shuffled hints do not (Table~\ref{tab:hint_shuffle}, App.~\ref{app:controls}); (P2) the sign-aware rule should beat symmetric and entropy-only weighting (Table~\ref{tab:design_controls}); and (P3) the interaction should be positive and largest on long-horizon tasks (Table~\ref{tab:wo_POPE}).
All three are borne out in Sec.~\ref{sec:experiments}.

\begin{table}[htbp]
\centering
\caption{
Results on six benchmarks ($\uparrow$: higher is better).
Open-source rows are mean $\pm$ std over four runs
(training seeds for fine-tuned models; bootstrap over examples for frozen baselines);
closed-source rows are single-run.
Bold: best among training strategies on the Qwen2.5-VL-7B backbone.
}
\label{tab:chair}
\footnotesize
\renewcommand{\arraystretch}{0.98}

\textbf{Hallucination Evaluations}

\vspace{0.3em}

\begin{tabular*}{\linewidth}{@{\extracolsep{\fill}}lccc@{}}
\toprule
\textbf{Model (Method)}
& \textbf{POPE $\uparrow$}
& \textbf{VideoHallucer $\uparrow$}
& \textbf{HalluBench $\uparrow$} \\
\midrule

\multicolumn{4}{c}{\textit{Closed-source Models}} \\
GPT-4o
& 86.9 & 53.3 & 55.0 \\
Gemini-1.5-Pro
& -- & 37.8 & 45.6 \\
\midrule

\multicolumn{4}{c}{\textit{Open-source Multimodal Base Models}} \\
LLaVA-OneVision-7B
& 69.8 $\pm$ 1.5
& 44.6 $\pm$ 1.2
& 61.6 $\pm$ 1.1 \\

InternVL2.5-8B
& 82.1 $\pm$ 0.6
& 50.5 $\pm$ 1.3
& 55.6 $\pm$ 0.7 \\

LLaMA3.2-11B
& 74.8 $\pm$ 1.3
& --
& 53.9 $\pm$ 1.1 \\
\midrule

\multicolumn{4}{c}{\textit{Open-source Multimodal Reasoning Models}} \\
R1-OneVision-7B
& 83.1 $\pm$ 0.8
& 42.4 $\pm$ 1.5
& 60.3 $\pm$ 1.0 \\

R1-VL-7B
& 82.4 $\pm$ 0.9
& 43.6 $\pm$ 1.4
& 62.7 $\pm$ 1.1 \\

Vision-R1-7B
& 81.1 $\pm$ 1.0
& 44.2 $\pm$ 1.2
& 58.3 $\pm$ 1.1 \\

Video-R1-7B
& 85.2 $\pm$ 0.7
& 45.1 $\pm$ 1.1
& 61.9 $\pm$ 0.8 \\
\midrule

\multicolumn{4}{c}{\textit{Qwen Based Models}} \\
Qwen2.5-VL-7B
& 81.3 $\pm$ 0.9
& 52.2 $\pm$ 1.1
& 64.1 $\pm$ 0.7 \\

+ SFT
& 82.2 $\pm$ 0.6
& 55.3 $\pm$ 0.9
& 67.9 $\pm$ 0.6 \\

+ GRPO
& 83.7 $\pm$ 0.8
& 52.0 $\pm$ 1.3
& 68.6 $\pm$ 0.9 \\

+ DAPO
& 84.5 $\pm$ 0.7
& 55.1 $\pm$ 1.0
& 66.1 $\pm$ 0.8 \\

+ PAPO
& 83.5 $\pm$ 0.8
& 54.1 $\pm$ 1.1
& 68.9 $\pm$ 0.7 \\

\rowcolor{blue!10}
+ DEEPO
& \textbf{85.8 $\pm$ 0.4}
& \textbf{57.6 $\pm$ 0.8}
& \textbf{69.7 $\pm$ 0.6} \\
\bottomrule
\end{tabular*}

\vspace{0.7em}

\textbf{General Reasoning}

\vspace{0.3em}

\begin{tabular*}{\linewidth}{@{\extracolsep{\fill}}lccc@{}}
\toprule
\textbf{Model (Method)}
& \textbf{MMBench $\uparrow$}
& \textbf{MMSTAR $\uparrow$}
& \textbf{VideoMMMU $\uparrow$} \\
\midrule

\multicolumn{4}{c}{\textit{Closed-source Models}} \\
GPT-4o
& 82.1 & 61.6 & 61.2 \\
Gemini-1.5-Pro
& 73.9 & 59.1 & 53.9 \\
\midrule

\multicolumn{4}{c}{\textit{Open-source Multimodal Base Models}} \\
LLaVA-OneVision-7B
& 81.7 $\pm$ 0.6
& 52.1 $\pm$ 1.0
& 31.2 $\pm$ 1.7 \\

InternVL2.5-8B
& 84.6 $\pm$ 0.4
& 55.3 $\pm$ 0.9
& 44.2 $\pm$ 1.4 \\

LLaMA3.2-11B
& 72.1 $\pm$ 0.8
& 46.7 $\pm$ 1.0
& -- \\
\midrule

\multicolumn{4}{c}{\textit{Open-source Multimodal Reasoning Models}} \\
R1-OneVision-7B
& 82.3 $\pm$ 0.5
& 47.4 $\pm$ 1.2
& 44.1 $\pm$ 1.4 \\

R1-VL-7B
& 73.8 $\pm$ 0.7
& 41.0 $\pm$ 1.3
& 42.7 $\pm$ 1.5 \\

Vision-R1-7B
& 83.8 $\pm$ 0.6
& 39.4 $\pm$ 1.4
& 39.7 $\pm$ 1.6 \\

Video-R1-7B
& 86.0 $\pm$ 0.5
& 54.0 $\pm$ 0.9
& 48.1 $\pm$ 1.2 \\
\midrule

\multicolumn{4}{c}{\textit{Qwen Based Models}} \\
Qwen2.5-VL-7B
& 83.1 $\pm$ 0.6
& 53.9 $\pm$ 0.8
& 43.9 $\pm$ 1.2 \\

+ SFT
& 82.9 $\pm$ 0.5
& 55.9 $\pm$ 0.7
& 45.8 $\pm$ 1.0 \\

+ GRPO
& 84.5 $\pm$ 0.6
& 57.8 $\pm$ 0.8
& 44.7 $\pm$ 1.3 \\

+ DAPO
& 84.4 $\pm$ 0.5
& 55.9 $\pm$ 0.9
& 48.7 $\pm$ 1.1 \\

+ PAPO
& 84.8 $\pm$ 0.5
& 58.1 $\pm$ 0.8
& 48.9 $\pm$ 1.0 \\

\rowcolor{blue!10}
+ DEEPO
& \textbf{86.4 $\pm$ 0.3}
& \textbf{60.8 $\pm$ 0.6}
& \textbf{53.3 $\pm$ 0.9} \\
\bottomrule
\end{tabular*}

\end{table}

\vspace{-0.8em}

\begin{table}[htbp]
\centering
\caption{
Ablation of the two branches and their interaction ($\uparrow$: higher is better).
All variants use Qwen2.5-VL-7B; all rows are four-seed means.
``w/o Expert Hint'' removes the whole signal-repair branch;
``w/o Gradient Scaling'' removes only the delivery branch, keeping the
entropy-triggered prefixes, prefix regularizer, and dropout.
The last row reports the interaction
$I = S_{\mathrm{full}} - S_{\mathrm{hint}} - S_{\mathrm{scaling}} + S_{\mathrm{GRPO}}$.
The 95\% CIs are $[-0.4,1.2]$/$[-1.3,1.7]$/$[-0.3,3.1]$/
$[-0.2,1.0]$/$[-0.8,1.2]$/$[1.1,6.9]$ (last bold);
only VideoMMMU excludes zero.
}
\label{tab:wo_POPE}
\footnotesize
\renewcommand{\arraystretch}{1}

\textbf{Hallucination Evaluations}

\vspace{0.3em}

\begin{tabular*}{\linewidth}{@{\extracolsep{\fill}}lccc@{}}
\toprule
\textbf{Model (Method)}
& \textbf{POPE $\uparrow$}
& \textbf{VideoHallucer $\uparrow$}
& \textbf{HalluBench $\uparrow$} \\
\midrule

Qwen2.5-VL-7B
& 81.3 & 52.2 & 64.1 \\

+ GRPO
& 83.7 & 52.0 & 68.6 \\

+ w/o Expert Hint
& 84.4 & 54.2 & 68.7 \\

+ w/o Gradient Scaling
& 84.7 & 55.2 & 68.2 \\

\rowcolor{blue!10}
+ DEEPO
& \textbf{85.8}
& \textbf{57.6}
& \textbf{69.7} \\
\midrule

Interaction $I$
& $+0.4$
& $+0.2$
& $+1.4$ \\
\bottomrule
\end{tabular*}

\vspace{0.7em}

\textbf{General Reasoning}

\vspace{0.3em}

\begin{tabular*}{\linewidth}{@{\extracolsep{\fill}}lccc@{}}
\toprule
\textbf{Model (Method)}
& \textbf{MMBench $\uparrow$}
& \textbf{MMSTAR $\uparrow$}
& \textbf{VideoMMMU $\uparrow$} \\
\midrule

Qwen2.5-VL-7B
& 83.1 & 53.9 & 43.9 \\

+ GRPO
& 84.5 & 57.8 & 44.7 \\

+ w/o Expert Hint
& 85.0 & 59.8 & 44.7 \\

+ w/o Gradient Scaling
& 85.5 & 58.6 & 49.3 \\

\rowcolor{blue!10}
+ DEEPO
& \textbf{86.4}
& \textbf{60.8}
& \textbf{53.3} \\
\midrule

Interaction $I$
& $+0.4$
& $+0.2$
& \textbf{$+4.0$} \\
\bottomrule
\end{tabular*}

\end{table}

\section{Experiments}
\label{sec:experiments}

\subsection{Experiment Setup}
\label{Experiment:Setup}

\noindent\textbf{Baselines, datasets, and protocol.}
We compare with closed-source models (GPT-4o~\cite{gpt4o2024}, Gemini-1.5-Pro~\cite{gemini152024}), open-source MLLMs (LLaVA-OneVision~\cite{llavaonevision2024}, InternVL2.5~\cite{internvl252024}, LLaMA3.2~\cite{llama322024}), recent multimodal reasoning models (R1-OneVision~\cite{r1onevision2025}, R1-VL~\cite{r1vl2025}, Vision-R1~\cite{visionr12025}, Video-R1~\cite{videor12025}), and, on the Qwen2.5-VL-7B backbone~\cite{qwen25vl2025}, the training strategies SFT, GRPO~\cite{grpoorigin2024}, DAPO~\cite{yu2025dapo}, PAPO~\cite{papo2025}, and DEEPO.
We train on the Video-R1 multimodal RL dataset~\cite{videor12025} and evaluate on POPE~\cite{li2023evaluating}, VideoHallucer~\cite{jiang2024videohallucer}, HallusionBench~\cite{guan2024hallusionbench} (hallucination) and MMBench~\cite{liu2023mmbench}, MMSTAR~\cite{liu2024mmstar}, VideoMMMU~\cite{li2024videommmu} (general reasoning).
Fine-tuned models are means over four training seeds with greedy decoding; frozen baselines are evaluated once with bootstrap standard deviations (Table~\ref{tab:chair}); the full implementation specification is in App.~\ref{app:training_details}.

\subsection{Performance}
\label{sec:performance}

Table~\ref{tab:chair} reports results on three hallucination-oriented and three general reasoning benchmarks.
DEEPO consistently achieves highly competitive performance across all benchmarks, setting new state-of-the-art results among the evaluated Qwen-based training strategies and outperforming GRPO, DAPO, and PAPO under the same backbone (85.8/57.6/69.7 on hallucination evaluations; 86.4/60.8/53.3 on general reasoning)---reducing hallucination does not trade off reasoning accuracy; the margin over GRPO's four-seed mean exceeds one seed standard deviation on five of six benchmarks (largest on VideoMMMU, $+8.6$), and the per-seed protocol is summarized in Table~\ref{tab:chair}.

\subsection{Ablation and Interaction}
\label{sec:ablation}

Table~\ref{tab:wo_POPE} ablates the two branches and reports their interaction.
\emph{The hint branch drives the reasoning-heavy gains} (\textit{w/o Gradient Scaling} improves five of six benchmarks, most sharply on VideoMMMU, $+4.6$); \emph{the scaling branch adds complementary gains} (\textit{w/o Expert Hint} improves five of six and is flat on VideoMMMU; added on top of hints it lifts VideoMMMU from $49.3$ to $53.3$).
The interaction is positive in the four-seed means on all six benchmarks but statistically significant only on long-horizon VideoMMMU; elsewhere the combination is consistent with additive gains.

\begin{table}[t]
\centering
\caption{
Delivery-branch design controls and trigger ablation (four-seed means, Qwen2.5-VL-7B); row 5 uses a rate-matched random trigger.
}
\label{tab:design_controls}
\small
\begin{tabular}{@{}lccc@{}}
\toprule
Variant & POPE $\uparrow$ & HallusionBench $\uparrow$ & VideoMMMU $\uparrow$ \\
\midrule
GRPO                              & 83.7 & 68.6 & 44.7 \\
w/ symmetric-$\tau$ weights         & 85.0 & 69.2 & 50.4 \\
w/ symmetric-inverse weights        & 85.2 & 69.3 & 51.1 \\
w/ sign-only weights                & 84.9 & 69.0 & 50.0 \\
w/ EMPG weights                     & 85.3 & 69.4 & 51.5 \\
w/ random trigger                   & 85.1 & 69.3 & 51.4 \\
\midrule
DEEPO (sign-aware, entropy trigger) & \textbf{85.8} & \textbf{69.7} & \textbf{53.3} \\
\bottomrule
\end{tabular}
\end{table}

\begin{table}[t]
\centering
\caption{
Mechanism diagnosis on the training distribution (four-seed pools): (a) all-wrong group fraction (\%) and mean group reward by $H_s$ decile (GRPO baseline); (b) hint activation on triggered, originally all-wrong queries; (c) baseline grouping ablation; (d) equal-compute and equal-supervision baselines.
All rows in (c)--(d) run the identical rollout pipeline of Algorithm~\ref{alg:deepo} ($G{=}8$ unhinted rollouts plus one hint-conditioned continuation per triggered query); the ``mixed'' row is the default model itself, not a rerun, and in the ``separated'' variant the unhinted group is normalized within itself while the singleton hinted continuation is scored against a batch-level EMA baseline over hint continuations---so no advantage is zeroed by construction and only the counterfactual penalty channel is removed.
}
\label{tab:mechanism_diag}
\small

\begin{tabular}{l*{10}{c}}
\toprule
$H_s$ decile & 1 & 2 & 3 & 4 & 5 & 6 & 7 & 8 & 9 & 10 \\
\midrule
All-wrong (\%) & 12 & 16 & 21 & 27 & 33 & 39 & 45 & 51 & 57 & 63 \\
Mean reward & 0.76 & 0.69 & 0.61 & 0.53 & 0.46 & 0.40 & 0.34 & 0.29 & 0.24 & 0.19 \\
\bottomrule
\end{tabular}

\vspace{2pt}

\begin{tabular}{lc}
\toprule
Metric & Value \\
\midrule
Converted to mixed groups ($\ge 1$ correct), \% & 60.0 \\
Failed-rollout mean advantage, pre / post & $0.000$ / $-0.212$ \\
Within-group advantage variance, pre / post & $0.000$ / $0.600$ \\
\bottomrule
\end{tabular}

\vspace{2pt}

\begin{tabular}{lccc}
\toprule
Baseline & POPE $\uparrow$ & HallusionBench $\uparrow$ & VideoMMMU $\uparrow$ \\
\midrule
Mixed (shared, default) & \textbf{85.8} & \textbf{69.7} & \textbf{53.3} \\
Separated (per-condition) & 84.4 & 68.5 & 49.7 \\
\bottomrule
\end{tabular}

\vspace{2pt}

\begin{tabular}{lccc}
\toprule
Variant & POPE $\uparrow$ & HallusionBench $\uparrow$ & VideoMMMU $\uparrow$ \\
\midrule
GRPO + extra budget      & 84.2 & 68.9 & 46.3 \\
SFT + GRPO (same traces) & 85.0 & 69.4 & 51.2 \\
\bottomrule
\end{tabular}
\end{table}

\subsection{Controls and Mechanism Evidence}
\label{sec:controls}
\label{sec:causal_controls}

\paragraph{Design controls.}
Replacing the sign-aware weight with any symmetric or sign-only alternative degrades all three benchmarks, as does replacing the entropy trigger with a rate-matched random one (Table~\ref{tab:design_controls}); likewise, entropy-triggered hints improve over the base method but trail full DEEPO (Table~\ref{tab:hint_control}, App.~\ref{app:controls}).
Conditioning the weight on the advantage sign---not entropy alone---is what delivers the correction, and the trigger gates the intervention to where it is needed.
The margin over EMPG's entropy-only factor is modest on POPE ($+0.5$) and largest on long-horizon VideoMMMU ($+1.8$); the R\'enyi-2 identity behind both rules is standard (Prop.~\ref{th:relation}), so the added value lies in the sign-aware allocation and the rollout-stage trigger, not in the identity itself.

\paragraph{Mechanism diagnosis.}
The rollout-stage diagnosis of Sec.~\ref{sec:expert_hint} is quantified in Table~\ref{tab:mechanism_diag}(a,b): all-wrong groups rise from $12\%$ to $63\%$ across $H_s$ deciles, and hints convert $60\%$ of triggered all-wrong groups into mixed ones.
Removing the shared baseline---with the identical rollout pipeline, and the singleton hinted continuation scored against a batch-level EMA baseline so that no advantage is zeroed by construction---degrades all three tracked benchmarks (Table~\ref{tab:mechanism_diag}(c), whose mixed row is the default model itself), isolating the counterfactual penalty on unhinted failures.

\paragraph{Fairness controls.}
The gains are not explained by extra compute or expert data (Table~\ref{tab:mechanism_diag}(d)): matching wall-clock with more GRPO steps, or matching traces with offline SFT before GRPO, both trail DEEPO, with the residual gap concentrated on VideoMMMU ($+2.1$)---offline SFT force-fits full trajectories and shifts the policy off its on-policy manifold, whereas DEEPO distills the same traces smoothly inside a single RL stage, avoiding the exploration collapse typical of multi-stage SFT-then-RL pipelines.
\emph{Paired analysis.} Across the four seeds, DEEPO leads SFT+GRPO on 4/4 seeds on VideoMMMU (paired bootstrap 95\% CI $[+1.2, +3.0]$) and on 3/4 seeds on POPE and HallusionBench (CIs $[-0.2, +1.4]$, $[-0.1, +0.9]$).
Further controls (App.~\ref{app:controls}, App.~\ref{app:further_controls}): shuffled hints show no distinct advantage over GRPO; \textsc{Deepo}-self retains $\approx 71\%$ of the VideoMMMU gain without expert data---consistent with the direct-grounding channel dominating, though source, quality, and coverage also differ between the variants, so the residual gap bounds rather than isolates the collapsed-group channel; amplified updates are enriched $1.75$--$1.80\times$ on span-annotated hallucinated content at the weight level ($1.15$--$1.25\times$ at the effective-step level); results transfer to InternVL2.5-8B ($85.0$ vs.\ $83.4$ GRPO on POPE).

\section{Conclusion}
\label{sec:conclusion}

We studied hallucination in RL of MLLMs as two weak points of the correction chain---on hard queries, unanimously wrong groups collapse the group-relative advantage to zero, and confident errors are gradient-invisible---and proposed DEEPO, a dual-stage enhancement pairing entropy-triggered expert prefixes at rollout time with advantage-sign-aware R\'enyi gradient preconditioning at optimization time.
Both branches improve over GRPO individually; their interaction is significant on long-horizon VideoMMMU and additive elsewhere; and the controls of Sec.~\ref{sec:experiments} isolate grounding content as the active ingredient, with stable training at $+18\%$ per-step cost over GRPO (App.~\ref{app:training_details}).
The guarantee currently concerns a local sample-wise surrogate rather than the global expected return.

{\small
\bibliographystyle{iclr2027_conference}
\bibliography{iclr2027_conference}
}

\newpage
\appendix
\onecolumn
\section{Additional Experimental Results}
\label{app:additional_experiments}

\subsection{Gradient-Mass Pilot}
\label{app:gradient_pilot}

\begin{figure*}[htbp]
    \centering
    \includegraphics[width=\linewidth]{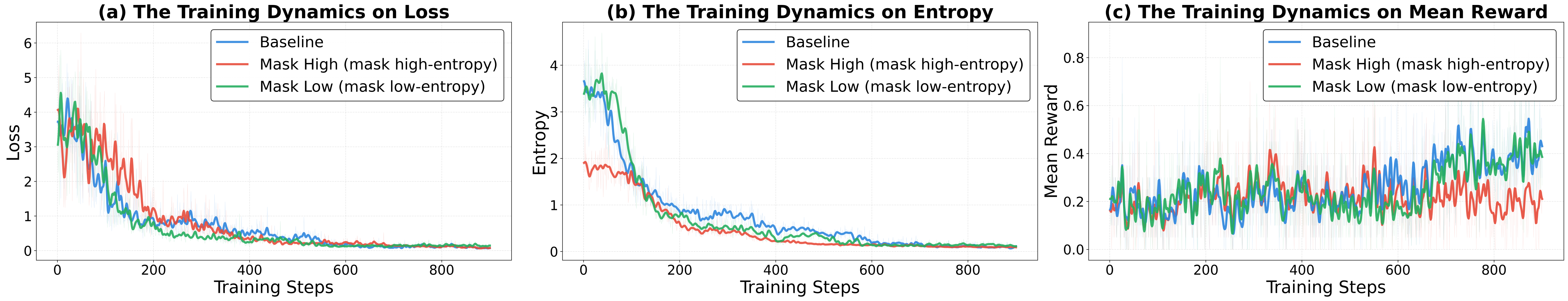}
      \caption{Effect of entropy-masked gradients. Masking low-entropy gradients preserves stable learning, while masking high-entropy gradients destabilizes training: high-entropy tokens carry the bulk of the update signal (Prop.~\ref{th:relation}); this pilot shows where gradient mass lives, not a hallucination outcome.}
    \label{fig:driver2-grad}
\end{figure*}

\subsection{Model Guess Analysis}
\label{app:model_guess_analysis}

To further examine uncertainty-driven guessing, we transform multiple-choice questions into cloze-style fill-in-the-blank queries, removing predefined options to compel models to answer from visual-semantic grounding rather than option cues.
This setting exposes semantic instability in free-form generation.

\begin{figure*}[htbp]
    \centering
    \includegraphics[width=\linewidth]{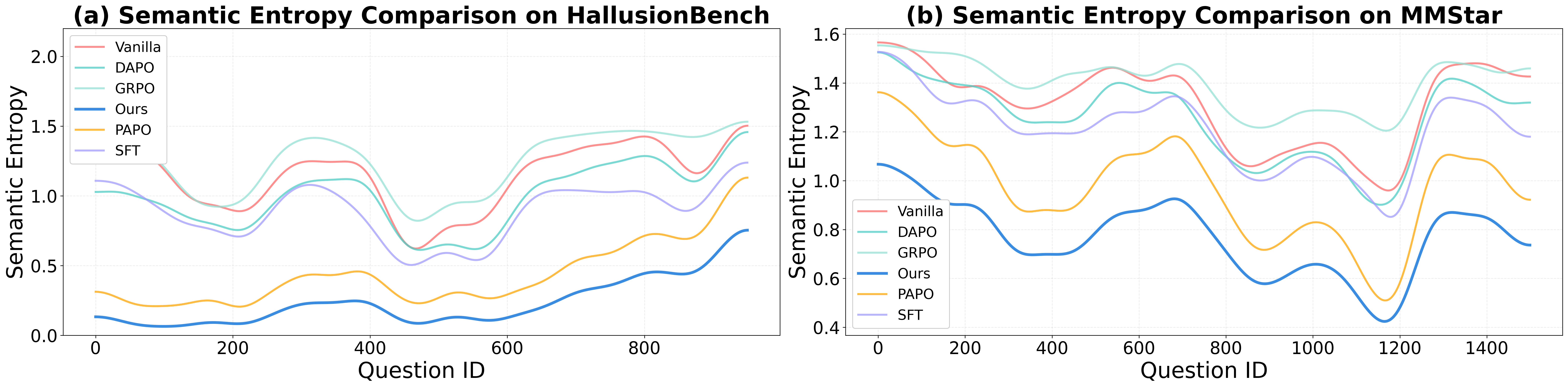}
    \caption{
    Semantic entropy on HallusionBench and MMStar (cloze-style conversion; questions sorted by the base model's $H_s$ for readability).
    Compared with the untrained base model, DAPO yields higher semantic entropy, while GRPO and PAPO reduce entropy but remain more variable.
    DEEPO maintains consistently lower and less dispersed semantic entropy, indicating reduced uncertainty-driven guessing.
    }
    \label{fig:training-guess}
\end{figure*}

As shown in Fig.~\ref{fig:training-guess}, DEEPO consistently produces the lowest semantic entropy across both HallusionBench and MMStar, with the smallest dispersion across questions (we operationalize ``more stable'' as the inter-question standard deviation of $H_s$).
In contrast, the base model, DAPO, GRPO, SFT, and PAPO exhibit higher or more variable entropy, indicating less stable answer-level commitments under ambiguous multimodal evidence.
We stress that lower $H_s$ is evidence of reduced guessing only when accuracy is preserved---a confidently wrong model would also have low entropy---so this analysis should be read together with the accuracy numbers in Table~\ref{tab:chair}.
These results support our motivation that semantic-entropy-triggered expert guidance reduces inconsistent semantic predictions and suppresses uncertainty-driven guessing during RL fine-tuning.

\subsection{Hyperparameter Sensitivity}
\label{app:hyperparameter_sensitivity}

We briefly check sensitivity to the initial semantic-entropy threshold $\taus$ (Table~\ref{tab:tau_sensitivity}) and the rollout budget $N$ (Table~\ref{tab:n_sensitivity}).
Varying $\taus$ over $\{0.2, 0.4, 0.5, 0.8, 0.9\}$ moves results by at most $0.7$ points on POPE and $0.5$ on VideoMMMU (four-seed means), as the EMA-updated threshold self-calibrates to the batch.
Increasing $N$ (co-varied with the GRPO group size $G$, so the joint effect of estimation budget and group size) improves results up to $N{=}8$--$12$ and degrades noticeably only in the small-$N$ regime ($-4.0$ on VideoMMMU at $N{=}4$); the default $N{=}8$ is a cost-driven choice.

\begin{table}[htbp]
\centering
\caption{
Sensitivity to the initial semantic entropy threshold $\taus$ (four-seed means).
}
\label{tab:tau_sensitivity}
\small
\setlength{\tabcolsep}{4pt}
\begin{tabular}{lcccccc}
\toprule
Initial $\taus$
& POPE $\uparrow$
& VideoHallucer $\uparrow$
& HallusionBench $\uparrow$
& MMBench $\uparrow$
& MMSTAR $\uparrow$
& VideoMMMU $\uparrow$ \\
\midrule
0.2 & 85.5 & 55.2 & 69.4 & 86.2 & 60.5 & 52.8 \\
0.4 & 85.7 & 55.7 & 69.5 & \textbf{86.5} & 60.7 & 53.1 \\
0.5 & 85.6 & 55.3 & 69.4 & 86.3 & 60.5 & 52.9 \\
0.8 & \textbf{85.8} & \textbf{57.6} & \textbf{69.7} & 86.4 & \textbf{60.8} & \textbf{53.3} \\
0.9 & 85.1 & 55.8 & \textbf{69.7} & \textbf{86.5} & 60.4 & 53.2 \\
\bottomrule
\end{tabular}
\end{table}

\begin{table}[htbp]
\centering
\caption{
Sensitivity to the rollout budget $N$ (co-varied with the GRPO group size $G$; four-seed means).
}
\label{tab:n_sensitivity}
\small
\setlength{\tabcolsep}{4pt}
\begin{tabular}{lcccccc}
\toprule
Rollout $N$
& POPE $\uparrow$
& VideoHallucer $\uparrow$
& HallusionBench $\uparrow$
& MMBench $\uparrow$
& MMSTAR $\uparrow$
& VideoMMMU $\uparrow$ \\
\midrule
4 & 84.7 & 54.2 & 68.5 & 85.4 & 59.1 & 49.3 \\
6 & 85.4 & 55.1 & 69.2 & 86.1 & 60.1 & 51.9 \\
8 & 85.8 & \textbf{57.6} & \textbf{69.7} & 86.4 & \textbf{60.8} & 53.3 \\
12 & \textbf{86.0} & 55.5 & 69.6 & \textbf{86.5} & 60.7 & \textbf{53.5} \\
\bottomrule
\end{tabular}
\end{table}

\subsection{Causal Controls and Hallucination-Targeted Baselines}
\label{app:controls}

This subsection accompanies Sec.~\ref{sec:causal_controls} and collects the causal controls.
Control runs (Table~\ref{tab:hint_shuffle}, Table~\ref{tab:deepo_self}, and Table~\ref{tab:mechanism_diag}(c) in the main text) follow the same four-seed protocol as the main results and are reported as means.

\paragraph{Grouping ablation.}
Table~\ref{tab:mechanism_diag}(c) (main text) compares the default mixed grouping against separated per-condition baselines.
Both variants run the identical rollout pipeline of Algorithm~\ref{alg:deepo}: $G{=}8$ first-pass unhinted rollouts plus one hint-conditioned continuation per triggered query, with matched compute and the same prefix regularizer.
They differ only in advantage computation: mixed pooling normalizes all nine samples jointly, whereas the separated variant normalizes the unhinted group within itself and scores the singleton hinted continuation against a batch-level EMA baseline over hint continuations of other triggered queries (so its advantage is well-defined and not identically zero, avoiding the size-1 group artifact).
The ``mixed'' row is the default DEEPO model itself, not a rerun, which is why its numbers coincide with Table~\ref{tab:chair}; the ``separated'' row is a separately trained run under this protocol.
The drop therefore isolates the loss of the counterfactual penalty on unhinted failures.

\begin{table}[h]
\centering
\caption{Controlled study on expert-hint supervision (four-seed means).
Entropy-triggered hints improve over the corresponding base method, yet remain below DEEPO---hint supervision alone does not explain the gains.
Hint rows use rollout conditioning only (no prefix regularizer, no dropout), unlike Table~\ref{tab:wo_POPE}'s ``w/o Gradient Scaling'' row.
}
\label{tab:hint_control}
\small
\setlength{\tabcolsep}{4pt}
\begin{tabular}{@{}lccccc@{}}
\toprule
Method
& Hint
& Trigger
& POPE $\uparrow$
& HallusionBench $\uparrow$
& VideoMMMU $\uparrow$ \\
\midrule
GRPO
& \xmark & \xmark
& 83.7 & 68.6 & 44.7 \\
GRPO + Expert Hint
& \cmark & \xmark
& 83.4 & 68.8 & 45.0 \\
GRPO + Entropy-triggered Hint
& \cmark & \cmark
& 84.2 & 69.0 & 46.7 \\
\midrule
DEEPO w/o Expert Hint
& \xmark & \xmark
& 84.4 & 68.7 & 44.7 \\
DEEPO
& \cmark & \cmark
& \textbf{85.8} & \textbf{69.7} & \textbf{53.3} \\
\bottomrule
\end{tabular}
\end{table}

\paragraph{Hint-shuffle control.}
Table~\ref{tab:hint_shuffle} repeats DEEPO training with the same entropy trigger but with prefixes drawn from \emph{mismatched} queries (shuffled within the batch). If the gain stemmed from generic regularization or longer context, shuffled hints would help equally.
The results confirm the prediction: shuffled hints show no distinct advantage over GRPO on any of the three benchmarks ($-0.5$/$-0.9$/$-0.2$; seed-to-seed variations overlap substantially), while correct hints improve by $+2.1$/$+1.1$/$+8.6$---grounding content, not the conditioning machinery, drives the gains.
\begin{table}[h]
\centering
\caption{Hint-shuffle control (four-seed means): entropy-triggered prefixes from mismatched queries. Shuffled hints should fail to help despite identical context length and trigger statistics.}
\label{tab:hint_shuffle}
\small
\begin{tabular}{lccc}
\toprule
Variant & POPE $\uparrow$ & HallusionBench $\uparrow$ & VideoMMMU $\uparrow$ \\
\midrule
GRPO (four-seed mean)                 & 83.7 & 68.6 & 44.7 \\
DEEPO (shuffled hints)                & 83.2 & 67.7 & 44.5 \\
DEEPO (correct hints, four-seed mean) & \textbf{85.8} & \textbf{69.7} & \textbf{53.3} \\
\bottomrule
\end{tabular}
\end{table}

\paragraph{Self-generated hints.}
Table~\ref{tab:deepo_self} evaluates \textsc{Deepo}-self, in which the hint for a triggered query is the prefix of the highest-reward verified-correct rollout from the same GRPO group (queries whose group contains no correct rollout receive no hint). This variant requires no dataset traces at all.
\textsc{Deepo}-self recovers most of the gain over GRPO ($+0.9$/$+0.0$/$+6.1$), trailing dataset-derived hints only modestly ($-1.2$/$-1.1$/$-2.5$).
Because a self-hint exists only when the group already contains a correct rollout, \textsc{Deepo}-self cannot exercise the advantage-reshaping channel on all-wrong groups; its retained gain (VideoMMMU: $6.1$ of $8.6$, $\approx 71\%$) indicates that the direct-grounding and supervision channel carries the bulk of the gain; since the two variants also differ in hint source, quality, and coverage, the residual gap is an upper bound on---not a clean isolation of---the collapsed-group coverage channel.
\begin{table}[h]
\centering
\caption{Self-generated hints (\textsc{Deepo}-self, four-seed means): prefixes come from the highest-reward verified-correct rollout within the same GRPO group; no dataset traces are used.}
\label{tab:deepo_self}
\small
\begin{tabular}{lccc}
\toprule
Variant & POPE $\uparrow$ & HallusionBench $\uparrow$ & VideoMMMU $\uparrow$ \\
\midrule
DEEPO (dataset hints, four-seed mean) & \textbf{85.8} & \textbf{69.7} & \textbf{53.3} \\
\textsc{Deepo}-self                   & 84.6 & 68.6 & 50.8 \\
\bottomrule
\end{tabular}
\end{table}

\subsection{Further Controls and Ablations}
\label{app:further_controls}

The following analyses isolate error localization (weight vs.\ effective step) and backbone transfer; the weighting-rule and trigger ablations are consolidated in Table~\ref{tab:design_controls} of the main text.

\paragraph{Error-localization analysis.}
Table~\ref{tab:error_localization} measures the relative per-token update intensity that the sign-aware rule assigns to hallucinated vs.\ non-error content on failed rollouts.

\emph{Annotation protocol.}
We randomly sampled 300 failed (reward-0) rollouts per benchmark (600 in total) from the evaluation generations of the GRPO baseline.
Two trained annotators independently marked hallucinated spans at the token level---visual entities, attributes, counts, and spatial or temporal relations not supported by the input image/video---following a written guideline that explicitly requires articles, connectives, and other function words to be labeled as non-error even when they are low-entropy.
Disagreements were adjudicated by a third annotator; inter-annotator agreement before adjudication was Cohen's $\kappa = 0.78$ (HallusionBench) and $0.81$ (VideoMMMU).
As an independent cross-check, GPT-4V was shown the same rollouts and visual inputs and asked to mark unsupported visual claims; it agreed with the adjudicated labels on $92\%$ of tokens on a $50\%$ subsample.
For each token we then recorded its scaling weight $g^\star$ under the DEEPO rule and its effective logit-step norm $\lVert\Delta z_t\rVert \propto w_t \lVert g_t\rVert$; the table reports both quantities on error vs.\ non-error tokens within failed rollouts, with 95\% bootstrap confidence intervals over rollouts (10{,}000 resamples).

\emph{Interpretation.}
Hallucinated spans are sparse in long-form generation: error tokens constitute $6.4\%$ (HallusionBench) and $7.1\%$ (VideoMMMU) of all tokens in the annotated failed rollouts; weighting by the enrichment factors, roughly $10.7\%$/$12.1\%$ of the weighted update mass lands on error tokens and the remaining $\sim\!88\%$ on non-error tokens---as it inevitably does for any sequence-level advantage formulation.
The table therefore supports a \emph{relative budget shift}---nearly doubled per-token update intensity on the error locus---not token-orthogonal deletion of hallucinated content.
Weight is also not gradient contribution: the effective-step enrichment ($1.15\times$/$1.25\times$) is attenuated relative to the weight enrichment because error tokens carry smaller raw score norms $\lVert g_t\rVert$---exactly the stall the compensation is designed to counteract (Thm.~\ref{thm:trace_asym}(iii)).
\begin{table}[h]
\centering
\caption{Error-localization analysis on failed rollouts: per-token update intensity on hallucinated vs.\ non-error content. Annotation: token-level hallucinated-span labeling by two human annotators with adjudication (600 rollouts; Cohen's $\kappa = 0.78$/$0.81$), cross-checked by GPT-4V ($92\%$ token agreement on a $50\%$ subsample). $\lVert\Delta z\rVert$: mean effective logit-step norm. CI: 95\% bootstrap over rollouts.}
\label{tab:error_localization}
\small
\begin{tabular}{lccccc}
\toprule
Benchmark & $g^\star$ (error) & $g^\star$ (non-error) & Enrichment $\uparrow$ & 95\% CI & $\lVert\Delta z\rVert$ (err / non-err) \\
\midrule
HallusionBench & 1.68 & 0.96 & 1.75 & $[1.58, 1.93]$ & 0.31 / 0.27 \\
VideoMMMU      & 1.80 & 1.00 & 1.80 & $[1.63, 1.98]$ & 0.35 / 0.28 \\
\bottomrule
\end{tabular}
\end{table}

\paragraph{Second backbone.}
Table~\ref{tab:backbone} reports generalization to InternVL2.5-8B.
\begin{table}[h]
\centering
\caption{Generalization to a second backbone (InternVL2.5-8B; four-seed means).}
\label{tab:backbone}
\small
\begin{tabular}{lccc}
\toprule
Variant & POPE $\uparrow$ & HallusionBench $\uparrow$ & VideoMMMU $\uparrow$ \\
\midrule
InternVL2.5-8B            & 82.1 & 55.6 & 44.2 \\
\quad + GRPO              & 83.4 & 57.0 & 45.3 \\
\quad + DEEPO             & 85.0 & 58.2 & 51.0 \\
\bottomrule
\end{tabular}
\end{table}



\section{Proof of the Entropy--Gradient Relationship}
\label{proof_entropy_gradient}

We prove Proposition~\ref{th:relation}. Let the policy over the $N$-token vocabulary be $\pi_j := \pi_\theta(y_j \mid q, y_{<t})$ with logits $z = (z_1,\dots,z_N)$, and let $g_k := \nabla_z \log \pi_k = e_k - \pi$ be the score-function gradient for sampled token $k$, where $e_k$ is the $k$-th standard basis vector and $\pi = (\pi_1,\dots,\pi_N)^\top$.
For each $k$,
\begin{equation}
\|g_k\|^2
= (1-\pi_k)^2 + \sum_{j\neq k} \pi_j^2
= 1 - 2\pi_k + \sum_{j=1}^{N} \pi_j^2 .
\end{equation}
Taking the expectation over $k \sim \pi$,
\begin{equation}
\mathbb{E}_{k\sim\pi}\big[\|g_k\|^2\big]
= \sum_{k=1}^{N} \pi_k \Big(1 - 2\pi_k + \sum_{j=1}^{N} \pi_j^2\Big)
= 1 - 2\sum_{k}\pi_k^2 + \sum_{j}\pi_j^2
= 1 - \sum_{j=1}^{N} \pi_j^2 .
\end{equation}
Since $H_2(\pi) = -\log \sum_j \pi_j^2$, we have $\sum_j \pi_j^2 = \exp(-H_2(\pi))$, and therefore
\begin{equation}
\mathbb{E}_{k\sim\pi}\big[\big\|\nabla_z \log \pi_k\big\|^2\big] = 1 - \exp\big(-H_2(\pi)\big),
\end{equation}
which is Eq.~\ref{eq:gradient_entropy_relation}. This is the trace of the categorical Fisher information matrix $F = \operatorname{diag}(\pi) - \pi\pi^\top$ in natural parameters, a standard identity (cf.\ \cite{amari2000methods}); equivalently, one minus Simpson's index of coincidence~\cite{simpson1949measurement}.

\section{Proof of Theorem~\ref{thm:trace_asym} and Remark~\ref{rem:signsym}}
\label{app:proof_trace_asym}

We first prove the sample-wise damped-proximal statement, then the sign-symmetry impossibility.

\paragraph{Setup and standing convention.}
Fix a training batch of \emph{already generated} trajectories.
Every advantage $A_t$ below is a fixed empirical scalar computed from the batch rewards (group-relative normalization included), and the sampled token $y_t$ at each position is fixed as well.
Accordingly, \emph{no expectation over $y_t \sim \pi_\theta$ is taken anywhere in this section with $A_t$ factored out}: the only distribution-level quantities are the Fisher matrix $F_t$ and its trace $\tau_t$, which are deterministic functions of the current policy at the state $s_t = (q, y_{<t})$ and do not depend on which token was sampled.
Write $\pi_j := \pi_\theta(y_j \mid s_t)$, $j = 1,\dots,N$, with logits $z = (z_1,\dots,z_N)$; for the (fixed) sampled token $y_t = y_k$ the score-function gradient with respect to the logits is $g_t = g_k = e_k - \pi$, where $e_k$ is the $k$-th standard basis vector.
The Fisher information matrix with respect to the logits is
\begin{equation}
F_t = \mathbb{E}_{y \sim \pi_\theta(\cdot \mid s_t)}\big[\, g_{y} g_{y}^\top \,\big] = \operatorname{diag}(\pi) - \pi\pi^\top,
\qquad
\tau_t := \Tr(F_t) = 1 - \textstyle\sum_j \pi_j^2 = 1 - \exp(-H_{2,t}),
\end{equation}
where the expectation defining $F_t$ involves the policy distribution only---it is the second moment of the score---and never the advantage $A_t$.

\paragraph{Part (i): sample-wise curvature bound and tangent minorant.}
$F_t \succeq 0$ since for any $u \in \mathbb{R}^N$, $u^\top F_t u = \sum_j \pi_j u_j^2 - \big(\sum_j \pi_j u_j\big)^2 = \operatorname{Var}_{J\sim\pi}[u_J] \ge 0$.
Consequently $\lambda_{\max}(F_t) \le \Tr(F_t) = \tau_t$, and for the fixed sampled token,
\begin{equation}
g_t^\top F_t g_t \;\le\; \lambda_{\max}(F_t)\,\lVert g_t\rVert^2 \;\le\; \tau_t \,\lVert g_t\rVert^2 .
\end{equation}
(For interpretation only, the distributional moments $\mathbb{E}_{y\sim\pi}\lVert g_y\rVert^2 = \tau_t$ (App.~\ref{proof_entropy_gradient}) and $\mathbb{E}_{y\sim\pi}[g_y^\top F_t g_y] = \Tr(F_t^2) = S_2 - 2S_3 + S_2^2 \le \tau_t^2$, with $S_r := \sum_j \pi_j^r$, identify $\tau_t$ as the expected gradient mass; the proof below needs only the sample-wise bound above, which holds for \emph{every} sampled token rather than in expectation over it.)
Restrict one-step logit updates to the score direction, $\Delta z_t = \eta\, w_t A_t g_t$ with $w_t \ge 0$.
The sample-wise second-order proximal surrogate with Levenberg--Marquardt damping $\epsz > 0$ evaluates to
\begin{equation}
Q_t(w_t)
\;=\; A_t\, g_t^\top \Delta z_t - \tfrac{\beta_{\mathrm{kl}}}{2}\, \Delta z_t^\top (F_t + \epsz I)\, \Delta z_t
\;=\; \eta A_t^2 \lVert g_t\rVert^2\, w_t - \frac{\beta_{\mathrm{kl}}\eta^2 A_t^2}{2}\,\big(g_t^\top F_t g_t + \epsz \lVert g_t\rVert^2\big)\, w_t^2 .
\end{equation}
Since $A_t$, $g_t$ are fixed sample quantities, $Q_t$ is an exact quadratic in $w_t$; the only approximation we introduce is the scalarization of the curvature term.
Define the trace-scalarized quadratic
\begin{equation}
\widetilde Q_t(w) = \eta A_t^2 \lVert g_t\rVert^2\, w - \frac{\beta_{\mathrm{kl}}\eta^2 A_t^2}{2}\,\lVert g_t\rVert^2\,(\tau_t + \epsz)\, w^2 .
\end{equation}
Since $g_t^\top F_t g_t + \epsz\lVert g_t\rVert^2 \le (\tau_t + \epsz)\lVert g_t\rVert^2$ by the bound above, $\widetilde Q_t(w) \le Q_t(w)$ for all $w \ge 0$, with $\widetilde Q_t(0) = Q_t(0) = 0$ and $\widetilde Q_t'(0) = Q_t'(0) = \eta A_t^2\lVert g_t\rVert^2$: $\widetilde Q_t$ is a tangent minorant of $Q_t$ on $\mathbb{R}_+$.

\paragraph{Part (ii): damped optimum, ascent, and cap.}
$\widetilde Q_t$ is concave in $w$ with unique maximizer
\begin{equation}
w_t^{\epsz} = \frac{\eta A_t^2 \lVert g_t\rVert^2}{\beta_{\mathrm{kl}}\,\eta^2 A_t^2 \lVert g_t\rVert^2 (\tau_t+\epsz)} = \frac{1}{\beta_{\mathrm{kl}}\,\eta\,(\tau_t + \epsz)} \;\le\; \frac{1}{\beta_{\mathrm{kl}}\,\eta\,\epsz} =: w_{\max},
\end{equation}
where the cap follows from $\tau_t \ge 0$; in particular the weight remains bounded as the curvature vanishes, unlike the undamped inverse-curvature rule $1/\tau_t$.
The achieved minorant value certifies strict sample-wise ascent:
\begin{equation}
Q_t(w_t^{\epsz}) \;\ge\; \widetilde Q_t(w_t^{\epsz}) = \frac{A_t^2\,\lVert g_t\rVert^2}{2\,\beta_{\mathrm{kl}}\,(\tau_t+\epsz)} \;>\; 0 = Q_t(0).
\end{equation}
The induced logit step itself stays controlled: $\lVert\Delta z_t\rVert = \eta\, w_t^{\epsz}\, |A_t|\,\lVert g_t\rVert = |A_t|\,\lVert g_t\rVert / \big(\beta_{\mathrm{kl}}(\tau_t+\epsz)\big)$, which vanishes smoothly as $g_t \to 0$ while its amplification \emph{relative to the unweighted rule} ($w_t = 1$) scales as $(\tau_t+\epsz)^{-1}$.
This is a conservative damped proximal step in logit space: the damping constant $\epsz$ caps the weight, and the guarantee degrades gracefully as $\tau_t \to 0$.

\paragraph{Part (iii): softmax-saturation regime.}
Let $\pi_y := \pi_\theta(y_t \mid s_t)$ denote the probability of the sampled token and $\delta_t := 1 - \pi_y$ the tail mass.
Using $\sum_{j \ne y}\pi_j^2 \le \big(\sum_{j \ne y}\pi_j\big)^2 = \delta_t^2$,
\begin{equation}
\lVert g_t\rVert^2 = (1-\pi_y)^2 + \sum_{j\ne y}\pi_j^2 \;\in\; \big[\,\delta_t^2,\; 2\,\delta_t^2\,\big],
\qquad
\tau_t = 1 - \pi_y^2 - \sum_{j\ne y}\pi_j^2 \;\in\; \big[\,2\delta_t(1-\delta_t),\; 2\delta_t - \delta_t^2\,\big].
\end{equation}
Hence, as the prediction saturates ($\pi_y \to 1$, i.e.\ $\delta_t \to 0$):
(a) the unweighted local slope vanishes \emph{quadratically}, $Q_t'(0) = \eta A_t^2 \lVert g_t\rVert^2 \le 2\eta A_t^2\,\delta_t^2$---this is the softmax-saturation stall;
(b) the compensated weight grows as $w_t^{\epsz} \ge \big(\beta_{\mathrm{kl}}\eta\,(2\delta_t - \delta_t^2 + \epsz)\big)^{-1} = \Omega(\delta_t^{-1})$ until it attains its cap $w_{\max}$;
(c) the certified ascent decays only \emph{linearly} in the tail mass,
\begin{equation}
Q_t(w_t^{\epsz}) \;\ge\; \frac{A_t^2\,\delta_t^2}{2\beta_{\mathrm{kl}}\,(2\delta_t - \delta_t^2 + \epsz)} \;=\; \Theta(\delta_t) \quad \text{for } \delta_t \gg \epsz .
\end{equation}
This is the formal content of saturation compensation: inverse-curvature weighting converts the local correction rate on a confident token from quadratic to first-order in the tail mass where the curvature dominates the damping ($\delta_t \gg \epsz$).
In the deep tail $\delta_t \ll \epsz$ the damping floor dominates: the weight saturates at $w_{\max} = (\beta_{\mathrm{kl}}\eta\epsz)^{-1}$ and the certified bound reverts to $O(\delta_t^2/\epsz)$---quadratic again, but with the $1/\epsz$ constant---while the induced step $\lVert\Delta z_t\rVert \le |A_t|\,\lVert g_t\rVert/(\beta_{\mathrm{kl}}\epsz)$ remains bounded and vanishes smoothly.
The theorem is thus a rate statement over the operational confidence band plus a boundedness statement in the tail; it does not eliminate saturation asymptotically.
This proves Theorem~\ref{thm:trace_asym}.

\paragraph{Sign-symmetry impossibility (Remark~\ref{rem:signsym}).}
Let $w_t = \nu(H_{2,t})$ be any advantage-sign-independent rule under a fixed batch budget $\sum_t w_t = C$.
Suppose there exist two samples $a,b$ with $H_{2,a} = H_{2,b}$ and $A_a < 0 < A_b$.
Then necessarily $w_a = \nu(H_{2,a}) = \nu(H_{2,b}) = w_b$, so no such rule can satisfy the correction-priority condition $w_a > w_b$.
More explicitly, define the corrective allocation objective
\begin{equation}
\mathcal{C}(w) := \sum_{t:\,A_t<0} |A_t|\, w_t\, \tau_t^{-1},
\end{equation}
measuring how much weighted update mass is assigned to negative-advantage samples after curvature compensation.
Starting from any symmetric rule, construct the budget-preserving perturbation $w_a' = w_a + \delta$, $w_b' = w_b - \delta$ with $0 < \delta < w_b$, leaving other weights unchanged.
The total budget is preserved, and since only sample $a$ contributes to $\mathcal{C}$ among this pair,
\begin{equation}
\mathcal{C}(w') - \mathcal{C}(w) = |A_a|\,(w_a' - w_a)\,\tau_a^{-1} = \delta\, |A_a|\, \tau_a^{-1} > 0 .
\end{equation}
So there always exists a budget-preserving asymmetric reallocation that strictly improves the corrective objective, while any symmetric rule $w_t = \nu(H_{2,t})$ remains invariant over equal-entropy pairs and cannot realize this preferential reallocation.

\paragraph{Scope and caveats.}
We record four qualifications on how Theorem~\ref{thm:trace_asym} should be read.
(a) The guarantee is \emph{sample-wise and local}: it concerns one damped proximal step in the logits of a single categorical conditional of one already-generated trajectory, with the batch advantages held fixed.
It is not a monotonic-improvement statement for the expected return under $y_t \sim \pi_\theta$, it makes no claim across re-sampled batches (over which the empirical $A_t$ are re-drawn), and it does not model shared parameters across token positions, PPO clipping, or Adam preconditioning---it is not a convergence or targeted-correction guarantee for the full network.
(b) $A_t$ enters $Q_t$ only through $A_t^2$: the theorem is sign-blind, so it justifies the \emph{magnitude} $1/(\tau_t+\epsz)$ of the corrective branch of Eq.~\ref{eq:phi_rule} but not its sign asymmetry---that asymmetry is a design choice motivated by Remark~\ref{rem:signsym}, and whether it outperforms symmetric rules is an empirical question, answered in Table~\ref{tab:design_controls}.
(c) The analysis treats a single token position with a local logit update; loss reweighting equals per-token logit stepping only to first order at the current iterate, and in practice $\beta_{\mathrm{kl}},\eta$ are absorbed by batch normalization, so only the \emph{shape} $1/(\tau_t+\epsz)$ survives---which is why we label Eq.~\ref{eq:phi_rule} a design rule rather than a corollary.
Finally, the KL penalty in Eq.~\ref{eq:grpo_total} is deliberately left unweighted: tokens receiving large corrective weights are permitted to move further from the reference policy, which we consider a feature of the correction branch rather than an oversight.

\section{Relation Analysis of Different Entropies}
\label{app:entropy_relation}

DEEPO uses two entropy signals at different levels, and it is natural to ask whether they measure the same thing.
They do not: semantic entropy $H_s$ captures answer-level dispersion across sampled rollouts (Sec.~\ref{sec:preliminaries}), while the R\'enyi-2 entropy $H_2$ characterizes token-level policy sharpness and, by Proposition~\ref{th:relation}, the expected score-gradient norm.
The two are linked through the token-level Shannon entropy
$H_1(\pi(\cdot \mid q, y_{<t})) = -\sum_{y\in\Vocab} \pi(y \mid q, y_{<t}) \log \pi(y \mid q, y_{<t})$,
the $\alpha \to 1$ limit of R\'enyi entropy, whose functional form mirrors that of $H_s$ (a Shannon entropy over meanings versus over tokens).
The chain
\begin{equation}
F(H_2;n)\;\ge\; H_1(p) \;\ge\; H_2(p)\;\ge\;\log n-\log\!\Big(1+2n\big(\log n-H_1(p)\big)\Big),
\end{equation}
summarized in the following proposition, makes the link quantitative.

\begin{proposition}[Entropy chain; known results]
\label{thm:entropy_chain}
Let $p=(p_1,\dots,p_n)$ denote the softmax distribution over an $n$-token vocabulary. Then the chain above holds, where
$F(H_2;n)=-\tfrac{1+(n-1)s}{n}\log\!\big(\tfrac{1+(n-1)s}{n}\big)-\tfrac{(n-1)(1-s)}{n}\log\!\big(\tfrac{1-s}{n}\big)$
and $s(H_2):=\sqrt{\tfrac{n e^{-H_2}-1}{n-1}} \in [0,1]$.
The upper bound is the sharp bound of \cite{harremoes2001bounds}, attained exactly by the two-level distribution
$p^\star=\big(\tfrac{1+(n-1)s}{n},\tfrac{1-s}{n},\dots,\tfrac{1-s}{n}\big)$;
$H_1 \ge H_2$ is R\'enyi monotonicity (equality iff $p$ is uniform on its support);
the lower bound follows from Pinsker's inequality and is tight only at the uniform distribution (it is strict, and in fact vacuous, at degenerate distributions---see Remark~\ref{rem:tightness}).
Self-contained derivations are given in App.~\ref{app:h1h2}.
\end{proposition}

The practical content for our design: the two branches do not require their entropy signals to be interchangeable or even strongly correlated---semantic entropy gates the rollout intervention, while token-level $H_2$ parameterizes optimization---and the chain above guarantees only that both are faithful measures of uncertainty at their respective levels.
\emph{A note on scope:} we deliberately do not use an empirical $H_s$--$H_1$ correlation analysis to justify combining the two signals; such a correlation would neither be necessary for the design nor sufficient to establish complementarity, and the $H_s$--$H_2$ chain is presented here solely to locate both quantities on a common uncertainty footing.

\section{Relation Between \texorpdfstring{$H_1$}{H1} and \texorpdfstring{$H_2$}{H2}}
\label{app:h1h2}

This appendix collects the relations between Shannon entropy $H_1$ and R\'enyi-2 entropy $H_2$ used in App.~\ref{app:entropy_relation}.
None of the results is new: $H_1 \ge H_2$ is R\'enyi monotonicity; the upper envelope $F(H_2;n)$ is the sharp entropy--index-of-coincidence bound of \cite{harremoes2001bounds}; the lower bound follows from Pinsker's inequality (cf.\ \cite{fedotov2003refinements}). We re-derive them in our notation for completeness.

Let $p=(p_1,\dots,p_n)$ be a discrete distribution on $n=|\Vocab|$ support points, $H_1(p) = -\sum_{i=1}^n p_i \log p_i$, $H_2(p) = -\log\big(\sum_{i=1}^n p_i^2\big)$, $S_2(p) = \sum_i p_i^2 = e^{-H_2}$.

\paragraph{($i$) $H_1 \ge H_2$.}
Let $X$ take value $p_i$ with probability $p_i$. Since $\log$ is concave, Jensen's inequality gives
$\mathbb{E}[\log X] \le \log \mathbb{E}[X]$, i.e.\ $\sum_i p_i \log p_i \le \log \sum_i p_i^2$; negating yields $H_1(p) \ge H_2(p)$, with equality iff $p$ is uniform on its support.

\paragraph{($ii$) Upper envelope $H_1 \le F(H_2;n)$~\cite{harremoes2001bounds}.}
Under the constraints $\sum_i p_i = 1$, $\sum_i p_i^2 = C$, the symmetric Schur-concave $H_1$ is maximized (by KKT / symmetry arguments) at a two-level distribution
\begin{equation}
p = (a,\underbrace{b,\dots,b}_{n-1\ \text{entries}}), \qquad b = \frac{1-a}{n-1}.
\end{equation}
The quadratic constraint gives $(n-1)C = na^2 - 2a + 1$; solving with $a \ge 1/n$ and introducing
\begin{equation}
s := \sqrt{\frac{nC-1}{n-1}} = \sqrt{\frac{n e^{-H_2}-1}{n-1}} \in [0,1],
\end{equation}
yields $a = \frac{1+(n-1)s}{n}$, $1-a = \frac{(n-1)(1-s)}{n}$, and substituting into $H_1(a,b) = -a\log a - (1-a)\log\frac{1-a}{n-1}$ gives the explicit envelope
\begin{equation}
F(H_2;n)
= -\frac{1+(n-1)s}{n}\log\!\Big(\frac{1+(n-1)s}{n}\Big)
-\frac{(n-1)(1-s)}{n}\log\!\Big(\frac{1-s}{n}\Big).
\end{equation}

\paragraph{($iii$) Lower bound via Pinsker.}
Let $u = (1/n,\dots,1/n)$. Since $\chi^2(p\|u) = n S_2(p) - 1 = n\|p-u\|_2^2$ and
\begin{equation}
D_{\mathrm{KL}}(p\|u) \ge \tfrac12\|p-u\|_1^2 \ge \tfrac12\|p-u\|_2^2 = \tfrac12\Big(S_2(p) - \tfrac1n\Big),
\end{equation}
we obtain $S_2(p) \le \big(1 + 2n D_{\mathrm{KL}}(p\|u)\big)/n$.
Taking $-\log$ and using $D_{\mathrm{KL}}(p\|u) = \log n - H_1(p)$:
\begin{equation}
H_2(p) \ge \log n - \log\!\Big(1 + 2n\big(\log n - H_1(p)\big)\Big).
\end{equation}

\begin{remark}[Tightness of the entropy chain]
\label{rem:tightness}
(i) At the uniform distribution $p = u$: $s = 0$, the envelope coincides with $u$, $F(\log n; n) = H_1 = H_2 = \log n$, and since $D_{\mathrm{KL}}(u\|u) = 0$ the lower bound holds with equality---the full chain is tight at the uniform point.
(ii) At any degenerate $p$: $s = 1$, the envelope coincides with $p$, so $F(0;n) = H_1 = H_2 = 0$ and the \emph{upper} bound is tight; the lower bound, however, evaluates to $\log n - \log(1 + 2n\log n) < 0$ for every $n \ge 2$ (since $1 + 2n\log n > n$), so it is \emph{strict}---indeed vacuous---there. More generally, the lower bound is negative whenever $D_{\mathrm{KL}}(p\|u) > (n-1)/(2n)$, and by the equality conditions of Pinsker's inequality it is attained only at $p = u$.
(iii) $H_1 = H_2$ iff $p$ is uniform on its support.
\end{remark}

\section{Protocol for the \texorpdfstring{$H_s$}{Hs}--\texorpdfstring{$H_1$}{H1} Alignment Analysis}
\label{correlation}

This appendix specifies the protocol for the empirical alignment analysis referenced in App.~\ref{app:entropy_relation}.

\paragraph{Quantities.}
For each evaluation question $q$ on HallusionBench and MMStar (multiple-choice converted to cloze-style, as in App.~\ref{app:model_guess_analysis}), we draw $M$ stochastic samples (temperature $1.2$) from the evaluated policy and compute:
(i) the semantic entropy $H_s(q)$ over answer-level clusters (Sec.~\ref{sec:preliminaries}, with the same protocol and budget $M$);
(ii) the token-level Shannon entropy $H_1$, averaged over answer-span tokens and over the $M$ samples,
\begin{equation}
\bar H_1(q) = \frac{1}{M}\sum_{i=1}^{M} \frac{1}{|y_i^{\mathrm{ans}}|} \sum_{t \in y_i^{\mathrm{ans}}} H_1\big(\pi(\cdot \mid q, y_{i,<t})\big),
\end{equation}
where $y_i^{\mathrm{ans}}$ denotes the extracted answer span of sample $i$.
Both quantities are computed at inference; expert hints are used only during \emph{training}, so ``with-hint'' columns below compare \emph{models trained with vs.\ without} the hint branch, not hint-conditioned decoding.

\paragraph{Statistics.}
We report the Pearson coefficient $\rho$ over questions between $\bar H_1(q)$ and $H_s(q)$, and the Spearman rank coefficient $\rho_{\mathrm{sp}}$ as a robustness check.
Significance uses a two-tailed $t$-test at level $\alpha = 0.01$ with statistic
$t = \rho\sqrt{M-2}/\sqrt{1-\rho^2}$,
where $M$ is the number of questions.

Table~\ref{tab:correlation} reports the resulting correlations ($\rho$: Pearson; $\rho_{\mathrm{sp}}$: Spearman; all $p<0.01$, two-tailed $t$-test).
The alignment is already present in the GRPO baseline and strengthens substantially with the hint branch on both benchmarks, supporting the claim that the two entropy signals are linked through predictive uncertainty rather than interchangeable.
\begin{table}[h]
\centering
\caption{Correlation between $\bar H_1$ and $H_s$ on multimodal benchmarks (Pearson $\rho$ / Spearman $\rho_{\mathrm{sp}}$), for models trained without and with the hint branch. All coefficients significant at $p<0.01$.}
\label{tab:correlation}
\small
\begin{tabular}{lcc}
\toprule
Model & HallusionBench $\rho$ / $\rho_{\mathrm{sp}}$ & MMStar $\rho$ / $\rho_{\mathrm{sp}}$ \\
\midrule
GRPO (w/o hint branch)   & 0.48 / 0.45 & 0.44 / 0.42 \\
DEEPO (w/ hint branch)   & \textbf{0.62} / \textbf{0.59} & \textbf{0.58} / \textbf{0.55} \\
\bottomrule
\end{tabular}
\end{table}

\section{Training Details}
\label{app:training_details}

\subsection{Implementation Specification}
\label{app:impl_spec}

For reproducibility we consolidate the design choices and hyperparameters that the main text references.
\emph{Semantic-entropy estimation:} answers are extracted from the rollout suffix after the final-answer marker, lowercased and stripped of articles and punctuation; numeric answers are canonicalized.
Answers are clustered into semantic-equivalence classes by pairwise DeBERTa-MNLI entailment (\texttt{microsoft/deberta-v2-xlarge-mnli}) under a lax rule---two answers are equivalent when they are mutually non-contradictory and at least one direction entails the other; cluster probabilities are count-based, $P(C_k\mid q)=|C_k|/N$, and $H_s$ is the natural-log Shannon entropy over these frequencies.
\emph{Prefix truncation:} the same final-answer marker delimits the thought/answer split $y^{*} = [\,y^{*}_{\mathrm{thought}}, y^{*}_{\mathrm{ans}}\,]$ of each expert trace; prefixes are cut from the thought segment only, at $L_h$ tokens given by Eq.~\ref{eq:prefix_len} with truncation cap $\alpha_{\max} = 0.5$, so the prefix never reaches the answer segment (we verified on the full training set that no prefix contains the gold answer string).
\emph{Semantic-leakage audit:} token-level exclusion does not by itself rule out semantic leakage, since an early reasoning prefix could still determine the conclusion.
We therefore audit every candidate prefix with the same DeBERTa-MNLI model used for clustering: the gold answer is verbalized as a declarative statement, and a prefix is used only if the model does not predict entailment from the prefix to that statement.
On the 20k training traces, $3.1\%$ of candidate prefixes fail this test and are discarded (the query then receives no prefix for that iteration); a manual inspection of 500 filtered prefixes found none that determines the final label.
\emph{Trigger:} the threshold is initialized at $\taus = 0.8$ and updated by exponential moving average with decay $\beta$; $H_s^{\max} = \log N$; hint dropout with rate $p_{\mathrm{drop}}$ is applied, and the trigger is evaluated before dropout.
\emph{Scaling:} $\tau_t$ is computed from $\pi_{\theta_{\mathrm{old}}}$'s fp32 log-probabilities with stop-gradient; $w_{\max} = 20$ ($\epsz = 0.05$); $g^\star$ is mean-normalized over the batch and applied after group advantage normalization and before clipping; we log the effective sample size of the weights and the clipped-token fraction as diagnostics.
\emph{Optimization:} clip $\epsclip = 0.2$, KL coefficient $\beta_{\mathrm{kl}} = 0.04$, prefix-loss weight $\lambda$, actor learning rate $5\times10^{-7}$; $G = 8$, sampling temperature $1.2$.



\begin{figure*}[h]
    \centering
    \includegraphics[width=\linewidth]{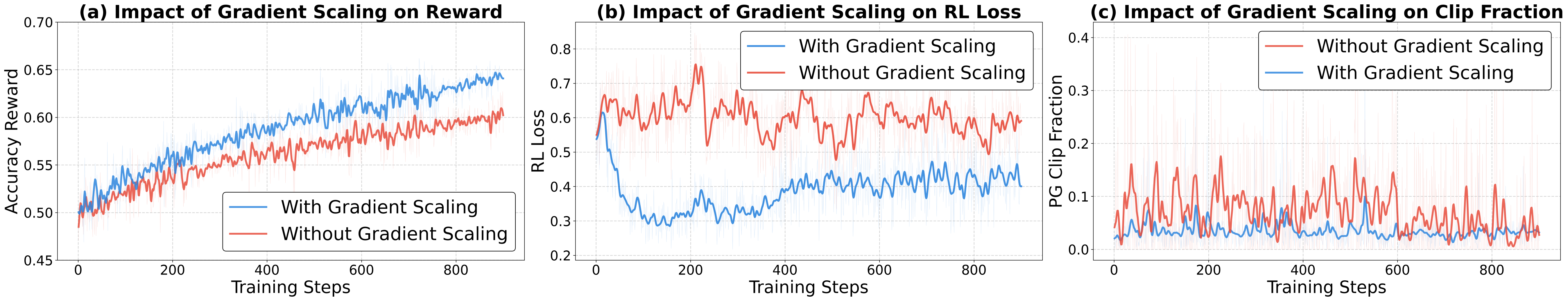}
      \caption{Effect of Gradient Scaling on RL stability.
Adaptive weighting stabilizes RL loss, accelerates reward convergence, and reduces policy clipping, yielding smoother and more robust training.}
    \vspace{-0.2em}

    
    \label{fig:training-stability}
\end{figure*}


We fine-tune the Qwen2.5-VL-7B model under the GRPO framework to enhance multimodal reasoning and reduce hallucination. The training covers both image--text and video--text modalities, unified through a shared tokenizer to maintain consistent encoding across modalities.

\subsection*{Data Preprocessing.}
We employ a unified visual--text tokenizer that handles both static and temporal visual inputs.
For image--text pairs, the images are dynamically resized based on token limits (4--16,384 tokens) while preserving aspect ratio.
For video--text pairs, the preprocessing dynamically adjusts frame counts between 4 and 64 based on original FPS and duration, ensuring each clip maintains a consistent spatial resolution and frame factor of 2.

\subsection*{Training Framework.}
We adopt the GRPO reinforcement learning setup~\cite{grpoorigin2024}, which is critic-free: advantages are estimated relative to the group of sampled responses (Sec.~\ref{sec:preliminaries}), and no value network is trained.
For each GRPO iteration, $G = 8$ candidate responses are sampled per multimodal input (image-- or video--question pair) using a temperature of 1.2.
The policy is updated with Adam at a learning rate of $5\times10^{-7}$.
A KL penalty of $0.04$ constrains the updated policy within the trust region defined by the reference model, stabilizing optimization.

\subsection*{Hyperparameters.}
Table~\ref{tab:hyperparameters} consolidates the configuration referenced throughout the paper.

\begin{table}[h]
\centering
\caption{Full hyperparameter configuration.}
\label{tab:hyperparameters}
\small
\begin{tabular}{ll}
\toprule
Parameter & Value \\
\midrule
Backbone & Qwen2.5-VL-7B~\cite{qwen25vl2025} \\
Training data & Video-R1~\cite{videor12025} multimodal RL split, 20k subset \\
GRPO group size $G$ & 8 \\
Rollout temperature & 1.2 \\
Optimizer (policy) & Adam, lr $= 5\times10^{-7}$ \\
Clip parameter $\epsclip$ & 0.2 \\
KL coefficient $\beta_{\mathrm{kl}}$ & 0.04 \\
Training steps & 2{,}000 \\
Evaluation cadence & every 100 steps, fixed held-out prompts \\
\midrule
SE estimation budget $N$ & 8 (${=}G$; co-varied with $G$, App.~\ref{app:hyperparameter_sensitivity}) \\
Initial SE threshold $\taus$ & 0.8 \\
Prefix truncation cap $\alpha_{\max}$ (Eq.~\ref{eq:prefix_len}) & 0.5 \\
Weight cap $w_{\max}$ ($\epsz = 1/w_{\max}$) & 20 \\
Global batch size (prompts / step) & 64 \\
EMA decay $\beta$ & 0.05 \\
Hint dropout rate $p_{\mathrm{drop}}$ & 0.20 \\
Prefix-loss weight $\lambda$ & 0.10 \\
Semantic-equivalence model & \texttt{microsoft/deberta-v2-xlarge-mnli} \\
Entailment rule & lax (mutual non-contradiction $+$ $\ge 1$ directional entailment) \\
Cosine threshold & N/A (not used) \\
Entropy estimator & cluster-frequency probability, natural-log entropy \\
\midrule
Hardware & 8 $\times$ NVIDIA A100 (80GB) \\
Wall-clock & ${\approx}3$ days (training only, excluding periodic evaluation) \\
\bottomrule
\end{tabular}
\end{table}

\subsection*{Dataset and Sampling.}
Our models are trained on the large-scale multimodal dataset introduced in Video-R1~\cite{videor12025}, selecting 20k samples as the final training subset.
Videos are preprocessed through the provided sampling pipeline, which adaptively determines frame counts based on original FPS and duration while maintaining aspect ratio and pixel limits.
Each sample is converted into an instruction-style multimodal prompt for consistency with the training objective.
The reward is verifiable: task accuracy on the final answer plus a format-compliance bonus, following~\cite{videor12025}.

\subsection*{Hardware and Schedule.}
All experiments are conducted on 8 $\times$ NVIDIA A100 (80GB) GPUs.
Training runs for 2{,}000 GRPO steps (approximately 3 days of pure training time at ${\approx}130$\,s/step for the full model), with evaluation conducted every 100 steps using a fixed set of held-out prompts; periodic evaluation and checkpointing add a small additional overhead not included in the per-step timing of Table~\ref{tab:efficiency}.

\subsection*{Per-Iteration Procedure}

Algorithm~\ref{alg:deepo} summarizes one training iteration of DEEPO. The procedure first estimates query-level uncertainty from grouped rollouts and uses an EMA threshold to identify ambiguous samples that benefit from expert guidance. For these samples, DEEPO conditionally injects a truncated expert prefix, while the token-level scaling branch independently reweights policy gradients according to local predictive uncertainty. Both branches are integrated into the same GRPO update, with hinted and unhinted trajectories sharing the group-relative baseline.

\begin{algorithm}[t]
\caption{One DEEPO training iteration}
\label{alg:deepo}
\begin{algorithmic}[1]
\REQUIRE policy $\pi_\theta$, reference $\pi_{\mathrm{ref}}$, a batch of queries with expert traces, group size $G$, threshold state $\hat\tau$ (initialized at $\taus$), EMA decay $\beta$, dropout rate $p_{\mathrm{drop}}$, cap $w_{\max}$, $\lambda$, $\beta_{\mathrm{kl}}$
\STATE Sample $G$ rollouts per query with $\pi_{\theta_{\mathrm{old}}}$ (temp.\ $1.2$)
\STATE Extract answers; cluster by DeBERTa-MNLI entailment (lax rule); compute $H_s(q)$ per query
\STATE Update threshold: $\hat\tau \leftarrow (1-\beta)\hat\tau + \beta\,\overline{H}_s$
\FORALL{queries $q$ with $H_s(q) > \hat\tau$}
   \STATE With prob.\ $1-p_{\mathrm{drop}}$: truncate the \emph{thought segment} of the expert trace to $L_h$ tokens (Eq.~\ref{eq:prefix_len}; answer segment $y^{*}_{\mathrm{ans}}$ excluded); sample continuation $y_q^{\mathrm{rem}} \sim \pi_{\theta_{\mathrm{old}}}(\cdot \mid q \oplus y_q^{\mathrm{pre}})$
\ENDFOR
\STATE Score all trajectories with the verifiable reward
\STATE Compute group-relative advantages over the full group: hinted continuations and first-pass unhinted rollouts share one baseline (in-group variance anchoring, Sec.~\ref{sec:expert_hint})
\STATE Compute $\tau_t = 1 - e^{-H_{2,t}}$ per token from $\pi_{\theta_{\mathrm{old}}}$ in fp32 (stop-grad); form $g^\star$ (Eq.~\ref{eq:phi_rule})
\STATE Update $\pi_\theta$ by minimizing Eq.~\ref{eq:grpo_total} (prefix masked in RL term; $\tilde A_t = g^\star A_t$ pre-clip)
\end{algorithmic}
\end{algorithm}

\subsection*{Efficiency Decomposition}

Table~\ref{tab:efficiency} decomposes the additional training cost introduced by the two DEEPO branches. Gradient scaling adds little overhead relative to GRPO, whereas the expert-hint branch accounts for most of the extra wall-clock time due to answer clustering and conditional continuation generation. 

\begin{table*}[t]
\centering
\caption{
Efficiency--performance trade-off. ``SE Time Share'' covers answer extraction and clustering for $H_s$ only; the conditional continuation pass and the prefix forward pass are separate delta components.
}
\label{tab:efficiency}
\small
\begin{tabular}{lcccc}
\toprule

Method 
& Step Time (s) $\downarrow$ 
& Wall-clock $\Delta$ 
& SE Time Share 
& MMBench $\uparrow$  \\
\midrule
Qwen2.5-VL-7B
&- &- &- &83.1 \\

+ GRPO
& \textbf{109.8 $\pm$ 2.7} 
& \textbf{1.00$\times$} 
& -- 
& 84.5 
 \\

+ DEEPO w/o Expert Hint \emph{(scaling only)} 
& 111.9 $\pm$ 2.9 
& +1.9\% 
& $\sim$0.0\% 
& 85.0 
 \\

+ DEEPO w/o Gradient Scaling \emph{(hint only)} 
& 123.6 $\pm$ 3.8 
& +12.6\% 
& 4.7\% $\pm$ 0.6\% 
& 85.5
\\

+ DEEPO full 
& 129.6 $\pm$ 4.1 
& +18.0\% 
& 5.4\% $\pm$ 0.7\% 
& \textbf{86.4}  \\
\bottomrule
\end{tabular}

\end{table*}

\end{document}